\documentclass[10pt,twocolumn,letterpaper]{article}
\PassOptionsToPackage{table}{xcolor}

\usepackage{iccv}      
\usepackage{multirow}
\usepackage{xcolor}  
\usepackage{float}

\definecolor{iccvblue}{rgb}{0.21,0.49,0.74}

\usepackage[pagebackref,breaklinks,colorlinks,allcolors=iccvblue]{hyperref}

\usepackage{tabularx}
\usepackage{multirow}
\usepackage{graphicx} 
\def\paperID{1727} 
\def\confName{ICCV}
\def\confYear{2025}

\title{Dual-Expert Consistency Model for Efficient and High-Quality Video Generation}

\author{
    {Zhengyao Lv}$^{2,3*}$ \quad {Chenyang Si}$^{1\ddag *}$ \quad {Tianlin Pan}$^{1,4}$ \quad {Zhaoxi Chen}$^{5}$ \\
    {Kwan-Yee K. Wong}$^{2}$ \quad {Yu Qiao}$^{3}$ \quad {Ziwei Liu}$^{5\dag}$ \\ \\
    $^1$Nanjing University  \quad $^2$The University of Hong Kong\quad  $^3$Shanghai Artificial Intelligence Laboratory \\ $^4$University of Chinese Academy of Sciences \quad $^5$S-Lab, Nanyang Technological University \\
    {\tt\small cszy98@gmail.com} \quad {\tt\small chenyang.si@nju.edu.cn} \quad {\tt\small pantianlin23@mails.ucas.ac.cn} \\  {\tt\small zhaoxi001@ntu.edu.sg} \quad {\tt\small kykwong@cs.hku.hk} \quad {\tt\small yu.qiao@siat.ac.cn} \quad {\tt\small ziwei.liu@ntu.edu.sg}
}

\begin{document}

\twocolumn[{
\renewcommand\twocolumn[1][]{#1}
\maketitle
\begin{center}
    \centering
    \vspace{-2.em}
    \includegraphics[width=.95\linewidth]{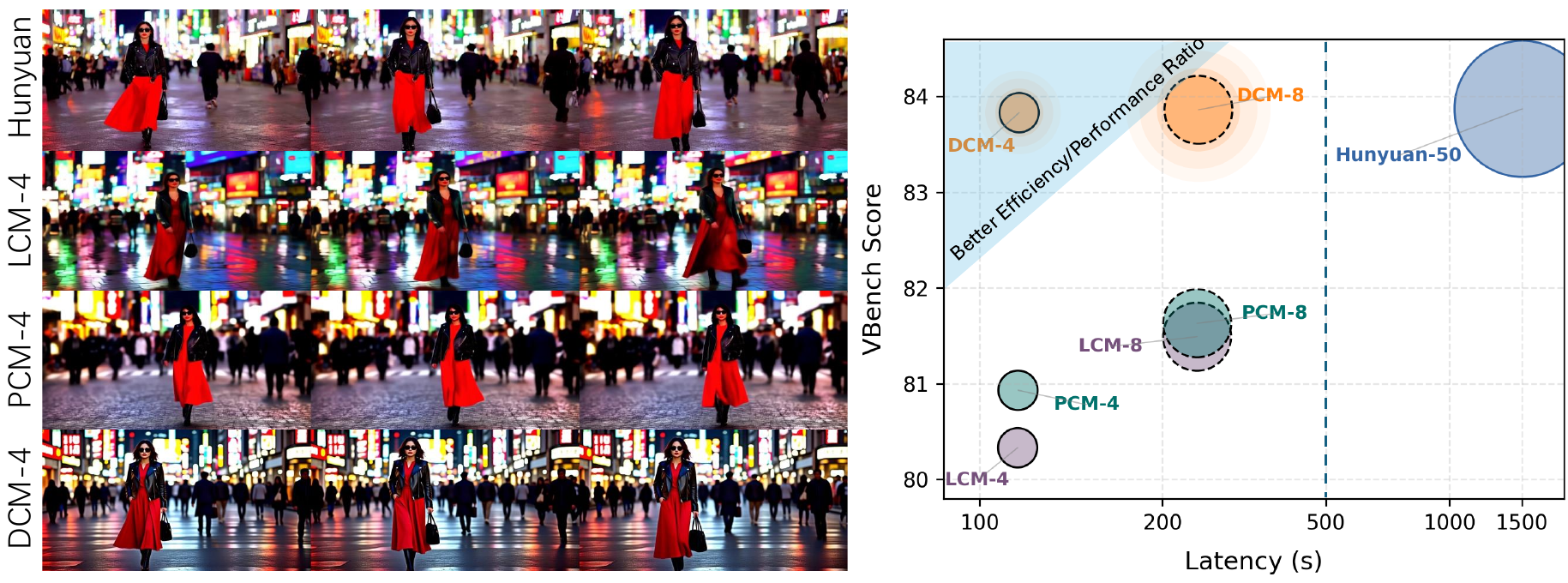}
    \vspace{-1.em}
    \captionof{figure}{
    Comparison of visual results between our DCM~(4 steps), the original HunyuanVideo, and other competing methods~(left). Comparison of latency and VBench score across different methods~(right).
    Latency is measured on two A100 GPUs under the video synthesis configuration of 129 frames at $1280\times720$ resolution.}
    \vspace{-0.3em}
    \label{fig:teaser}
\end{center}
}] 
\renewcommand{\thefootnote}{} 
\footnotetext{*Equal Contribution. $^{\ddag}$ Project Leader $^{\dag}$Corresponding Author.}

\vspace{-2em}
\begin{abstract}
Diffusion Models have achieved remarkable results in video synthesis but require iterative denoising steps, leading to substantial computational overhead. 
Consistency Models have made significant progress in accelerating diffusion models. 
However, directly applying them to video diffusion models often results in severe degradation of temporal consistency and appearance details. 
In this paper, by analyzing the training dynamics of Consistency Models, we identify a key conflicting learning dynamics during the distillation process: there is a significant discrepancy in the optimization gradients and loss contributions across different timesteps.
This discrepancy prevents the distilled student model from achieving an optimal state, leading to compromised temporal consistency and degraded appearance details.
To address this issue, we propose a parameter-efficient \textbf{Dual-Expert Consistency Model~(DCM)}, where a semantic expert focuses on learning semantic layout and motion, while a detail expert specializes in fine detail refinement.
Furthermore, we introduce Temporal Coherence Loss to improve motion consistency for the semantic expert and apply GAN and Feature Matching Loss to enhance the synthesis quality of the detail expert.
Our approach achieves state-of-the-art visual quality with significantly reduced sampling steps, demonstrating the effectiveness of expert specialization in video diffusion model distillation.
Our code and models are available at \href{https://github.com/Vchitect/DCM}{https://github.com/Vchitect/DCM}.
\end{abstract}
  
\vspace{-2em}
\section{Introduction}
Diffusion Models~\cite{ho2020denoising} have achieved remarkable progress in image and video synthesis~\cite{rombach2022high,peebles2023scalable,yang2024cogvideox}. However, they require multiple iterations to model the probability flow Ordinary Differential Equation (ODE)~\cite{song2020denoising} and rely on increasingly large denoising networks, resulting in substantial computational overhead that limits their practicality in real-world applications.

To mitigate this constraint, Consistency Distillation~\cite{song2023consistency} has emerged as an efficient knowledge distillation framework to reduce the sampling timesteps. 
It leverages a pre-trained diffusion model as the teacher and trains a student model to directly map any point along the ODE trajectory to the same solution, thereby ensuring the self-consistency property. 
Despite enabling few-step sampling, it often struggles with visual quality, particularly in challenging video synthesis, leading to distorted layouts, unnatural motion, and degraded details.

\begin{figure*}[t]
\centering
\includegraphics[width=.9\linewidth]{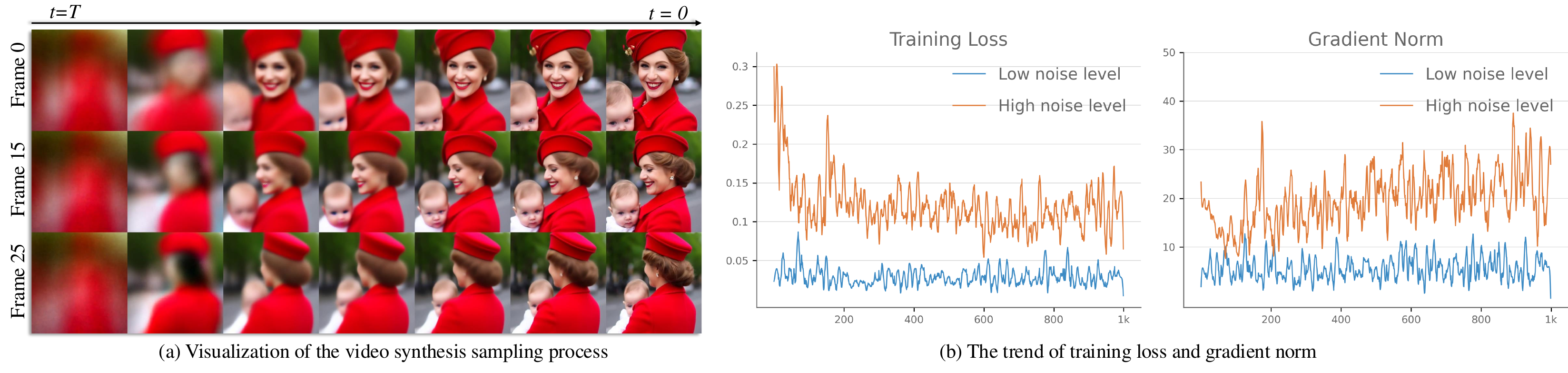}
\vspace{-1em}
\caption{Visualization of the video synthesis process and the trend of loss variation. (a) In the early stages of sampling, the results change significantly and rapidly, whereas in the later stages, the changes become gradual and smooth. (b) During distillation, the loss and gradient norm of the student model exhibit significant differences between samples with high and low noise levels.}
\vspace{-1.7em}
\label{fig:losstrend}
\end{figure*}

To ground this inherent issue, we analyzed the sampling dynamics of video diffusion models, as shown in Fig.~\ref{fig:losstrend}~(a). Our key observations are that \textbf{\textit{the differences between adjacent steps are substantial in the early stages of sampling, whereas the changes become more gradual in the later stages.}}
This discrepancy arises because the early steps primarily focus on synthesizing semantic layout and motion, while the later steps emphasize refining fine details.
These findings suggest that the student model may learn different patterns and exhibit distinct learning dynamics when trained on high-noise and low-noise samples.
We visualized the magnitude and gradient of the consistency loss during the distillation process and observed significant differences between high and low noise levels, as shown in Fig.~\ref{fig:losstrend}~(b).
This variation indicates that jointly distilling a single student model to capture both semantic layout and fine-detail synthesis may introduce optimization interference, potentially leading to suboptimal results.

To validate this assumption, we trained two expert denoisers.
We first divide the ODE trajectory of the pre-trained model into two phases: the semantic synthesis phase and the detail refinement phase.
We then train two distinct student expert denoisers, each responsible for fitting one of these sub-trajectories. During inference, we dynamically select the corresponding expert denoiser based on the noise level of samples to predict the next position in the ODE trajectory. The results demonstrate that the combination of the two student expert denoisers achieves better performance, thereby confirming the validity of our hypothesis.

However, this straightforward baseline involves training two student models which is not efficient enough.
To further enhance parameter efficiency, we analyze the parameter differences between the two expert denoisers and identify that the primary differences lie in \textbf{1)} embedding layers where the input parameters include timesteps, and \textbf{2)} the linear layers within the attention layers.
Based on this insight, we propose a parameter-efficient \textbf{Dual-Expert Consistency Model~(DCM)}. 
Specifically, we first train a semantic expert denoiser on the semantic synthesis trajectory. We then freeze this expert and introduce a new set of timestep-dependent layers, incorporating a LoRA~\cite{hu2021lora} into the linear layers of the attention blocks.
Subsequently, we fine-tune these newly added layers on the detail refinement trajectory. 
In this manner, we decouple the optimization of the two expert denoisers with minimal additional parameters and computational cost, achieving visual results comparable to those obtained with two separate experts.

Given the differing training dynamics of the semantic and detail expert denoisers, we introduce distinct optimization objectives beyond the original consistency loss.
To enforce temporal coherence in the semantic expert denoiser, we introduce a Temporal Coherence Loss, which guides it to capture motion variations across frames.
To enhance the fine-grained content synthesized by the detail expert denoiser, we introduce a generative adversarial~(GAN)~\cite{goodfellow2014generative} loss and incorporate a Feature Matching loss.
Specifically, we alternately optimize the student model and the discriminator in the feature space, encouraging the generator to synthesize visual content that aligns with the output distribution of the teacher model. The Feature Matching term enhances supervision over intermediate features, thereby stabilizing the GAN training.

Our proposed DCM accelerates sampling while preserving both semantic and detail quality, as shown in Fig.~\ref{fig:teaser}.
In summary, our contributions are as follows:
\begin{itemize}
\item We analyze the training dynamics of Consistency Models and identify a key conflict in the distillation process: discrepancies in loss contributions and optimization gradients across noise levels hinder optimal learning, leading to suboptimal visual quality.
\item We propose a parameter-efficient Dual-Expert Consistency Model that decouples the expert denoisers distillation, mitigating the conflict and improving visual quality with minimal parameter and computational cost.
\item To enhance visual quality, we introduce Temporal Coherence Loss for the semantic expert and GAN loss with Feature Matching term for the details expert, improving both temporal consistency and detail quality.
\end{itemize}
\section{Related Work}
\vspace{-0.5em}
\subsection{Diffusion Models For Video Synthesis}
\vspace{-0.5em}
\textbf{Video Diffusion Models} have witnessed rapid advancements with diffusion models~\cite{ho2020denoising,ho2022video,blattmann2023align}, driving a wide range of visual content generation applications\cite{zhang2023controlvideo,hu2024animate, zhang2025videoelevator,huang2025dreamphysics,zhang2025framepainter}.
Building on the Diffusion Transformer~(DiT)~\cite{peebles2023scalable} pre-training, a notable breakthrough is the development of high-fidelity video diffusion models~\cite{opensora,kuaishou,genmo2024mochi,ma2024latte,pku_yuan_lab_and_tuzhan_ai_etc_2024_10948109,hong2022cogvideo,yang2024cogvideox,kong2024hunyuanvideo}. However, scaling these models for long videos incurs significant training and inference costs. LTX-Video~\cite{HaCohen2024LTXVideo} designed Video-VAE
that achieves a high compression ratio for efficient self-attention. Pyramid flow~\cite{jin2024pyramidal} introduced a unified pyramidal flow matching algorithm for efficient video generative modeling. 
The acceleration method based on sparse attention~\cite{zhang2025fast, zhang2025faster} and feature cache~\cite{zhao2024real,lv2024fastercache} has also improved inference efficiency. 
Moreover, while efficient fast diffusion samplers~\cite{song2020denoising, liu2022pseudo,lu2022dpm,lu2022dpmpp,karras2022elucidating} reduce inference steps, further reduction often severely degrades performance. Diffusion distillation offers a promising way to reduce sampling steps while maintaining visual quality.

\subsection{Diffusion Model Distillation}
\vspace{-0.5em}
Diffusion distillation~\cite{luo2023comprehensive} aims to distill knowledge from pre-trained diffusion models to student models, reducing inference cost. Prior works can be generally classified into two categories based on their distillation mechanisms.

\noindent\textbf{Trajectory-preserving distillation} methods exploit the fact that diffusion
models learn an Ordinary Differential Equation (ODE)
trajectory and aim to predict the exact teacher output in fewer steps.
Among the earliest studies on diffusion distillation, Luhman et al.~\cite{luhman2021knowledge} and DSNO~\cite{zheng2023fast} proposed training the student model using noise-image pairs precomputed by the teacher model with an ODE solver.
Progressive distillation~\cite{salimans2022progressive,meng2023distillation} reduces the final number of sampling steps by iteratively applying the distillation process to halve the number of sampling steps of previous model.
Instaflow~\cite{liu2022flow,liu2023instaflow} progressively learns straighter flows, enabling accurate one-step predictions over larger distances.
Consistency models~\cite{song2023consistency,song2023improved,luo2023latent,luo2023lcm, liu2024scott, zheng2024trajectory,lu2024simplifying}, BOOT~\cite{gu2023boot} and TRACT~\cite{berthelot2023tract} learn to map samples along the ODE trajectory to another point to achieve self-consistency.
The consistency trajectory model~\cite{kim2023consistency} was designed to mitigate discretization inaccuracies and accumulated estimation errors in the multistep consistency model sampling.
PCMs~\cite{wang2024phased} phase the ODE trajectory into several sub-trajectories and only enforce the self-consistency property on each sub-trajectory, thus alleviating the limitations of CMs. 
TCM~\cite{lee2024truncated} generalizes consistency training to the truncated time range to prevents the truncated-time training from collapsing to a trivial solution.
Trajectory-preserving distillation enables stable optimization but can degrade visual quality, leading to blurriness or distortions when sampling with fewer steps.

\noindent\textbf{Distribution-matching distillation} methods bypass the ODE trajectory and aim to train the student model to generate samples whose distribution aligns with that of the teacher diffusion model. Some methods reduce the distribution gap between the student and teacher models through adversarial training~\cite{xiao2021tackling,luo2024you,kang2024distilling,sauer2024adversarial,sauer2024fast,lin2025diffusion,chen2024nitrofusion,xu2024ufogen}. 
Other methods~\cite{luo2023diff,luo2024one,yin2024one,yin2024improved,yin2024slow,zhou2024score} achieve diffusion distillation by score distillation~\cite{poole2022dreamfusion}. Notably, DMD~\cite{yin2024one} aligns the one-step generator with the distribution of teacher model by minimizing an approximate KL divergence, whose gradient is the difference between the target and synthetic distribution score functions. Recently works~\cite{lin2024sdxl,ren2024hyper,kohler2024imagine}
have also tried to integrate the advantages of trajectory-preserving and distribution-matching methods. Hyper-SD~\cite{ren2024hyper} introduced the trajectory-segmented consistency distillation and used DMD~\cite{yin2024one} for one-step generation enhancement.

Previous works have primarily focused on distilling image synthesis diffusion models, with some efforts~\cite{lin2024animatediff,wang2024animatelcm,zhai2024motion} extending to the distillation of small-scale video synthesis models~\cite{guo2023animatediff,wang2023modelscope}. However, these methods are limited to synthesizing low-resolution and short-sequence videos. Seaweed-APT~\cite{lin2025diffusion} proposed adversarial post-tuning against real data following diffusion pre-training for one-step high-resolution 2-second duration video generation. A recent work~\cite{yin2024slow} extended DMD~\cite{yin2024one} for video synthesis in four sampling steps. Due to the inherent complexity of video synthesis and the increasing model scale, research on diffusion distillation for video synthesis remains limited, and its performance is yet to be fully explored.
\section{Methodology}
\begin{figure*}[t]
\centering
\includegraphics[width=.925\linewidth]{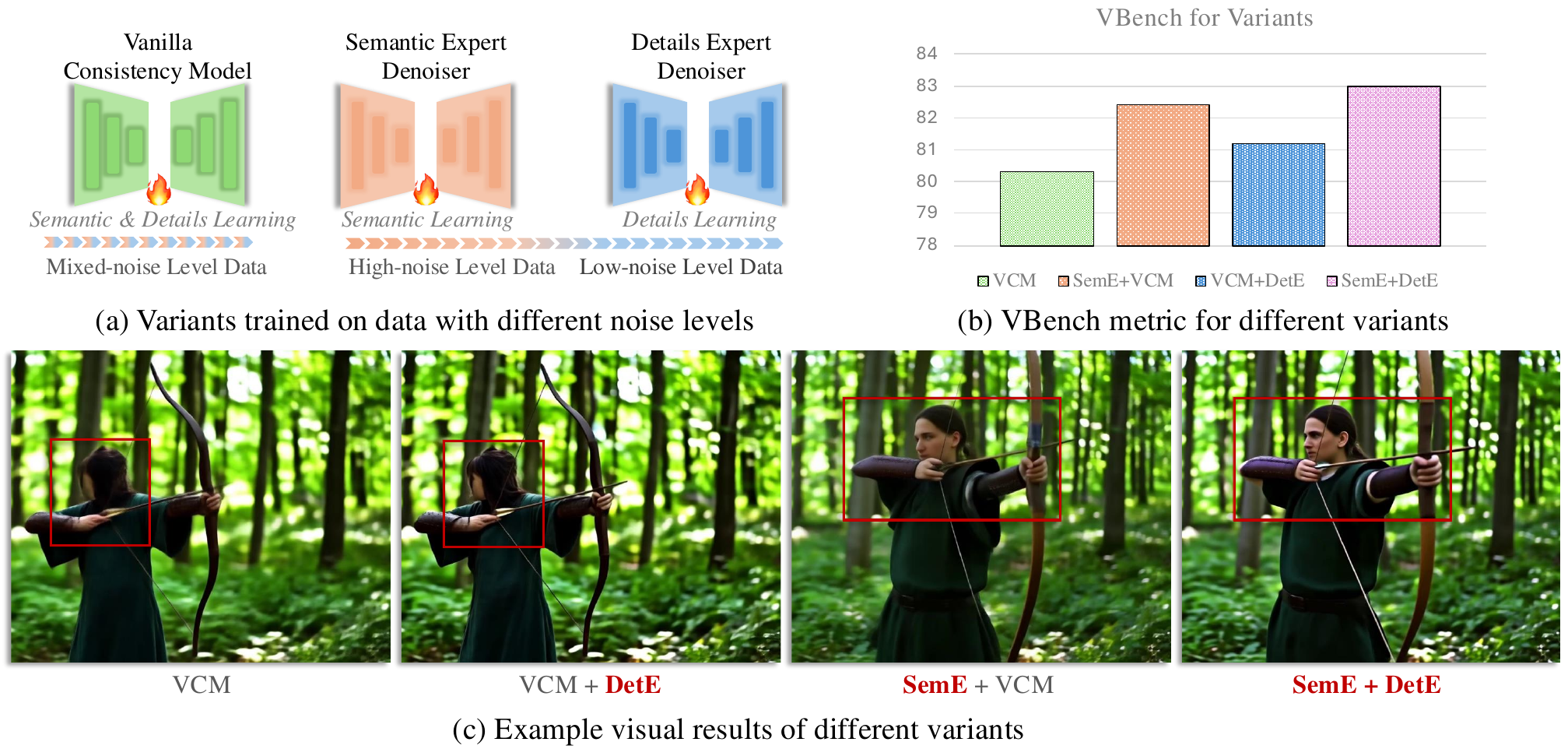}
\vspace{-1em}
\caption{Comparison of the visual quality of denoiser variants trained at different noise level samples. By optimizing two expert denoisers to decouple the distillation process into semantic learning and detail learning, and combining them during inference, we achieve the best quantitative and qualitative visual results.}
\vspace{-2em}
\label{fig:motivation}
\end{figure*}
\vspace{-0.3em}
\subsection{Preliminary}
\vspace{-0.3em}
\noindent\textbf{Diffusion Model} is a generative framework with a forward and reverse process.

In the forward process, noise is progressively added to clean data $\boldsymbol{x}_0 \sim p_{\mathrm{data}}(\boldsymbol{x}_0)$, degrading the signal:
\begin{align}
q(\boldsymbol{x}_t|\boldsymbol{x}_0) = \mathcal{N}(\boldsymbol{x}_t;\sqrt{\alpha_t}\boldsymbol{x}_0,\sqrt{1-\alpha_t}\boldsymbol{I}),
\label{eq:forward}
\end{align}
where $\{\alpha_t\}_{t=1}^T$ controls the noise schedule. 

The reverse process, typically parameterized by a UNet or transformer $\boldsymbol{\epsilon}_{\theta}$, is trained to predict the noise:
\begin{align}
\mathcal{L}_{DM} = \mathbb{E}_{\boldsymbol{x}, \boldsymbol{\epsilon} \sim \mathcal{N}(0,1),t} \big[||\boldsymbol{\epsilon} - \boldsymbol{\epsilon}_{\theta}(\boldsymbol{x}_t,t)||^2_2\big].
\end{align}
During inference, a clean sample $\boldsymbol{x}_0$ can be recovered through iterative denoising:
\begin{align}
p(\boldsymbol{x}_{t-1}|\boldsymbol{x}_t) = \mathcal{N}(\boldsymbol{x}_{t-1};\mu_{\theta}(\boldsymbol{x}_t,t),\Sigma_{\theta}(\boldsymbol{x}_t,t)),
\end{align}
where $\mu_{\theta}$ and $\Sigma_{\theta}$ are learned parameters.

\noindent\textbf{Consistency Distillation} utilizes a pre-trained model $\boldsymbol{\epsilon}_{\theta}$ as the teacher $F_T$ to distill its knowledge into a student model $F_S$ initialized with $\boldsymbol{\epsilon}_{\theta}$, allowing for faster sampling with fewer steps~\cite{song2023consistency}. 
Specifically, consistency distillation trains the student model $F_S$ to directly map any point $\boldsymbol{x}_{t_n}$ on the solution trajectory of the ODE solver $\Phi$ to its endpoint $\boldsymbol{x}_{t_{end}}$. The learning objective can be formulated as:
\begin{equation}
\begin{aligned}
\mathcal{L}_{CD} = \mathbb{E}_{\boldsymbol{x},t_n}||\Phi(\boldsymbol{x}_{t_{n}},F_S(\boldsymbol{x}_{t_{n}},t_n,c),t_{end})- \\ \Phi(\hat{\boldsymbol{x}}_{t_{n-1}},F_S^{-}(\hat{\boldsymbol{x}}_{t_{n-1}},t_{n-1},c),t_{end})||_2^2.
\end{aligned}
\end{equation}
Here, $F_S^-$ is the exponential moving average (EMA) of $F_S$ and $\hat{\boldsymbol{x}}_{t_{n-1}}$ is the next point on the ODE solution trajectory computed by the teacher model $F_T$:
\begin{equation}
\begin{aligned}
\hat{\boldsymbol{x}}_{t_{n-1}} = \Phi(\boldsymbol{x}_{t_{n}},F_T(\boldsymbol{x}_{t_{n}},t_n,c),t_{n-1}).
\end{aligned}
\end{equation}
Consistency distillation has garnered widespread attention and research~\cite{luo2023latent,ren2024hyper,wang2024phased} due to its ease of stable training. 

\begin{figure*}[t]
\centering
\includegraphics[width=0.9\linewidth]{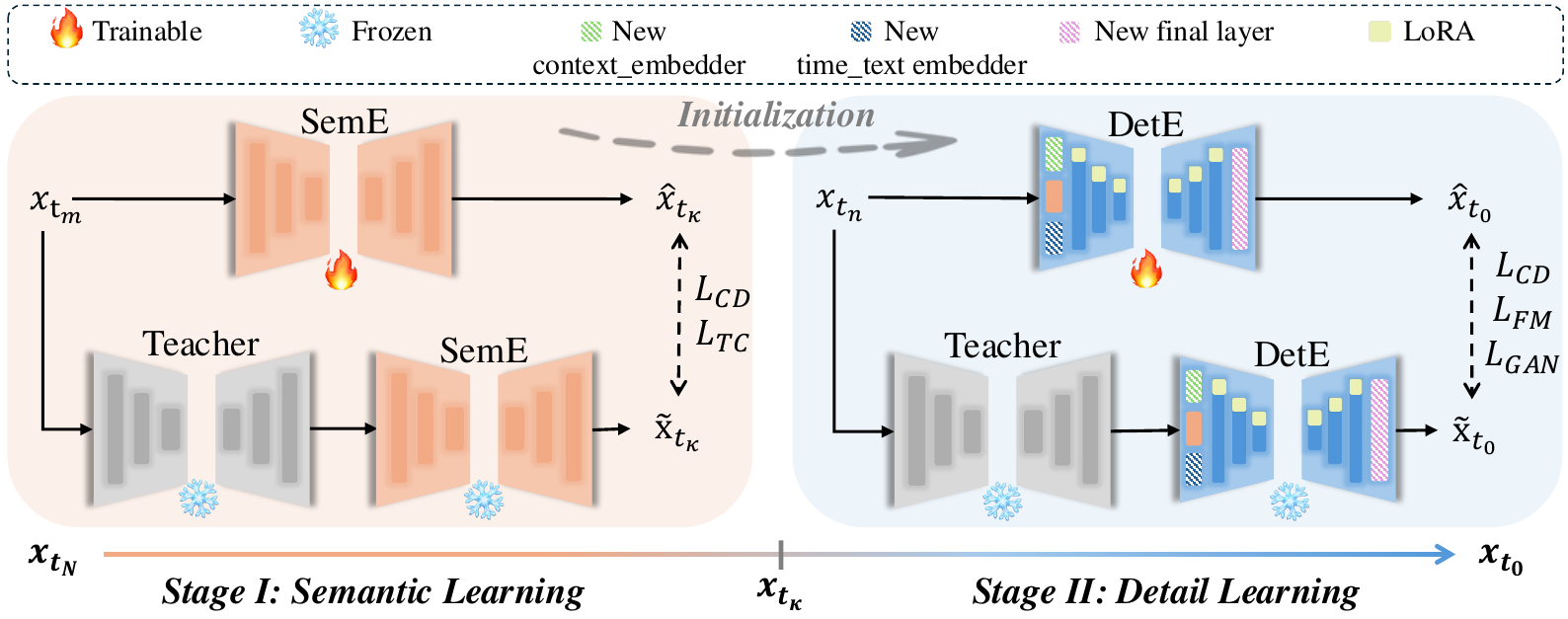}
\vspace{-0.4em}
\caption{The training process of DCM consists of two stages. In the semantic learning stage, we train SemE on high-noise samples with consistency loss and temporal coherence loss as the learning objectives. In the detail learning stage, we initialize DetE with the weights of SemE and introduce a set of time-dependent layers and LoRA. DetE is then trained on low-noise samples, where \textbf{only the newly added layers and LoRA are updated}. The learning objectives in this stage include consistency loss, GAN loss, and Feature Matching loss.}
\vspace{-1.0em}
\label{fig:method}
\end{figure*}

\begin{figure}[!t]
    \centering
    \includegraphics[width=\columnwidth]{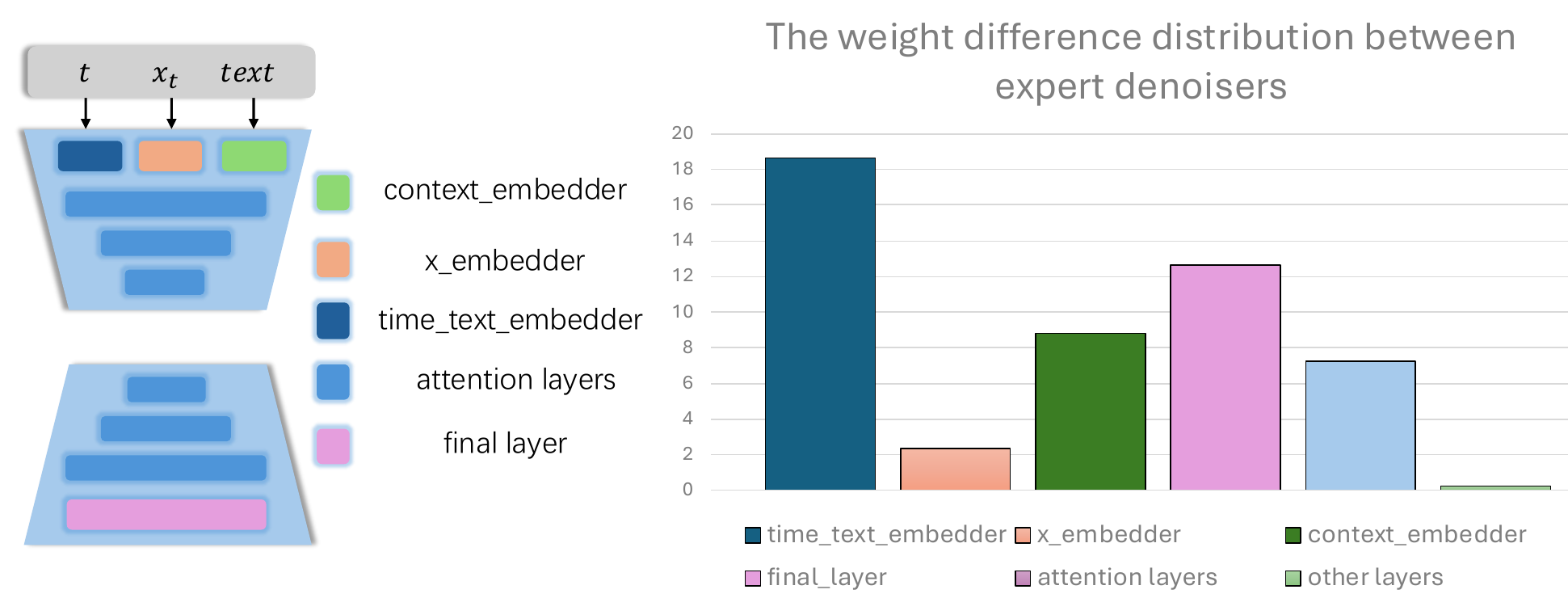}
    \caption{Weight difference distribution between expert denoisers. We employ the normalized L1 distance to quantify the difference between the weights.}
    \vspace{-1.75em}
    \label{fig:weight}
\end{figure}
\vspace{-0.4em}
\subsection{Suboptimal Solution in Consistency Distillation}
\vspace{-0.4em}
Although consistency distillation has demonstrated promising results in class-conditioned image synthesis and text-to-image model distillation, it falls short in more challenging video synthesis diffusion models, where issues arise such as distorted layouts, unnatural motion, and degraded details.

Due to the limited capacity of the student model, consistency distillation struggles to address these issues simultaneously.
By tracking the video synthesis process, we find that in the early stages of sampling, the sampling results vary significantly and rapidly, whereas in the later stages, the transitions become more gradual and smooth, as shown in Fig.~\ref{fig:losstrend}~(a).
In the early stages of sampling, rapid changes establish the semantic layout and motion of the video, while in the later stages, the model gradually refines the details with smooth adjustments.
These findings imply that the student model may learn different patterns with distinct training dynamics when distilled with high-noise and low-noise samples.
By visualizing the trends of consistency loss and gradient norm during distillation, we find significant differences in the student model when distilled with high- and low-noise samples, as shown in Fig.~\ref{fig:losstrend}~(b).
This suggests that jointly optimizing a student model for both semantic and fine-grained details synthesis may introduce inefficient optimization, constraining its fitting capacity and leading to suboptimal performance.

To validate our hypothesis, we conducted an experiment on the HunyuanVideo~\cite{kong2024hunyuanvideo} text-to-video diffusion model. Specifically, we divided the ODE solution trajectory ($\boldsymbol{x}_{N},\boldsymbol{x}_{N-1},...,\boldsymbol{x}_{1},\boldsymbol{x}_{0}$) of the pre-trained model into two sub-trajectories, using $t_{\kappa}$ as the boundary (we set $N=50$ and $\kappa=37$ by default). The first part ($\{\boldsymbol{x}_{t_{i}}\}_{i=\kappa}^{N}$) primarily focuses on synthesizing the semantic layout and motion, while the second part ($\{\boldsymbol{x}_{t_{j}}\}_{j=0}^{\kappa}$) emphasizes semantic refinement and high-quality detail generation. As shown in Fig~\ref{fig:motivation} (a), we optimized two distinct student models, semantic expert denoiser $F_{\text{SemE}}$ and details expert denoiser $F_{\text{DetE}}$, to fit each sub-trajectory:
\begin{equation}
\begin{aligned}
\mathcal{L}_{\text{SemE}} = \mathbb{E}_{\boldsymbol{x},t_m\in[t_{\kappa},t_{N}]}||\Phi(\boldsymbol{x}_{t_{m}},F_{\text{SemE}}(\boldsymbol{x}_{t_{m}},t_m,c),t_{\kappa})- \\ \Phi(\hat{\boldsymbol{x}}_{t_{m-1}},F_{\text{SemE}}^{-}(\hat{\boldsymbol{x}}_{t_{m-1}},t_{m-1},c),t_{\kappa})||_2^2,
\end{aligned}
\label{eq:S1}
\end{equation}
\begin{equation}
\begin{aligned}
\mathcal{L}_{\text{DetE}} = \mathbb{E}_{\boldsymbol{x},t_n\in [t_{0},t_{\kappa}]}||\Phi(\boldsymbol{x}_{t_{n}},F_{\text{DetE}}(\boldsymbol{x}_{t_{n}},t_n,c),t_{0})- \\ \Phi(\hat{\boldsymbol{x}}_{t_{n-1}},F_{\text{DetE}}^{-}(\hat{\boldsymbol{x}}_{t_{n-1}},t_{n-1},c),t_{0})||_2^2.
\end{aligned}
\label{eq:S2}
\end{equation}
Expert denoiser $F_{\text{SemE}}$ is optimized to synthesize coherent semantic layouts and motion, while expert denoiser $F_{\text{DetE}}$ learns to generate high-quality details.
During inference, we dynamically switch the expert denoiser based on the sampling stage.
To assess the impact of each expert denoiser and the effectiveness of decoupled training, we evaluate four variants:
\textbf{a)} VCM: The vanilla consistency model is used throughout the sampling process.
\textbf{b)} SemE + VCM: SemE is applied in the first sub-trajectory, transitioning to VCM in the second.
\textbf{c)} VCM + DetE: VCM is applied initially, followed by DetE.
\textbf{d)} SemE + DetE: SemE and DetE are integrated for the sampling process.
According to the Fig.~\ref{fig:motivation}~(b) and (c), SemE and DetE have respectively learned to model semantics and details, each outperforming VCM in their respective aspects. This validates our hypothesis, demonstrating that decoupled optimization is superior to jointly training a single model for both tasks. 

\vspace{-0.5em}
\subsection{Parameter-efficient Dual-Expert Distillation}
\vspace{-0.5em}
While training two expert denoisers improves video quality, it significantly increases model parameters and GPU memory consumption during inference.
Through the analysis of parameter similarity between the two expert denoisers, we found that the primary differences in model parameters lie in the embedding layers $\Psi$ where the input parameters include timesteps and the linear layers within the attention layers $\Lambda$, as illustrated in Fig.~\ref{fig:weight}.

Based on the above observations, we propose the parameter-efficient Dual-Expert distillation strategy, as illustrated in Fig.~\ref{fig:method}.
Specifically, our training scheme is divided into two stages.
\textbf{1)} Initialize the semantic expert denoiser $F_{\text{SemE}}$ with the teacher model $F_T$ and optimize all its parameters on the sub-trajectory $\{\boldsymbol{x}_{t_{i}}\}_{i=\kappa}^{N}$. 
\textbf{2)} Use the optimized $F_{\text{SemE}}$ as the initialization of $F_{\text{DetE}}$ and freeze it. Then add a new set of timestep-dependent embedding layers $\Psi$ and LoRA~\cite{hu2021lora} $\Lambda^\dag$ of attention blocks. Optimize the newly added parameters ($\Psi$ and $\Lambda^\dag$) on its sub-trajectory $\{\boldsymbol{x}_{t_{j}}\}_{j=0}^{\kappa}$. 
In this way, we significantly reduce the number of parameters required for decoupling the optimization process of semantic modeling and detail learning, with minimal computational cost, while maintaining the visual quality of the synthesized videos.

\vspace{-0.5em}
\subsection{Expert-specific Optimization Objective}
\vspace{-0.5em}
In addition to the consistency objectives mentioned in Eq.~\ref{eq:S1} and Eq.~\ref{eq:S2}, we also designed expert-specific optimization objectives for the semantic expert denoiser $F_{\text{SemE}}$ and details expert denoiser $F_{\text{DetE}}$, as shown in Fig.~\ref{fig:method}.

\noindent\textbf{Temporal Coherence Loss\quad}To enhance the temporal coherence in the video synthesized by the semantic expert denoiser $F_{\text{SemE}}$, we introduce the Temporal Coherence Loss $\mathcal{L}_{TC}$, which emphasizes and guides the $F_{\text{SemE}}$ to focus on the variations and motion at corresponding positions among different frames:
\begin{equation}
\begin{aligned}
\boldsymbol{x}_{t_\kappa} &= \Phi(\boldsymbol{x}_{t_{m}},F_{\text{SemE}}(\boldsymbol{x}_{t_{m}},t_m,c),t_{\kappa}), \\
\hat{\boldsymbol{x}}_{t_\kappa} &= \Phi(\hat{\boldsymbol{x}}_{t_{m-1}},F_{\text{SemE}}^{-}(\hat{\boldsymbol{x}}_{t_{m-1}},t_{m-1},c),t_{\kappa}), \\
\mathcal{L}_{TC} &= ||(\boldsymbol{x}^{t_\kappa}_{l:L}-\boldsymbol{x}^{t_\kappa}_{0:L-l}) - (\hat{\boldsymbol{x}}^{t_\kappa}_{l:L} -\hat{\boldsymbol{x}}^{t_\kappa}_{0:L-l})||^2_2.
\end{aligned}
\end{equation}
Here $\boldsymbol{x}^{t_\kappa}_{l:L}$ represents the video latents from the $l$-th to the $L$-th channel along the temporal axis. This temporal coherence loss encourages the semantic expert denoiser $F_{\text{SemE}}$ to preserve consistent motion and spatial relationships between frames, ensuring more fluid videos synthesis.

\noindent\textbf{Generative Adversarial Loss\quad}The effectiveness of the generative adversarial (GAN)~\cite{goodfellow2014generative} loss in high-quality detail synthesis has been validated in many distribution-matching distillation methods. 
We introduce the GAN loss into the training of the details expert denoiser and incorporate a Feature Matching loss to stabilize the training. We first obtain $\boldsymbol{x}_{t_0}$ and $\hat{\boldsymbol{x}}_{t_0}$ with the details expert denoiser $F_{\text{DetE}}$, teacher model $F_T$ and ODE solver $\Phi$: 
\begin{equation}
\begin{aligned}
\boldsymbol{x}_{t_0} &= \Phi(\boldsymbol{x}_{t_{n}},F_{\text{DetE}}(\boldsymbol{x}_{t_{n}},t_n,c),t_{0}), \\
\hat{\boldsymbol{x}}_{t_0} &= \Phi(\hat{\boldsymbol{x}}_{t_{n-1}},F_{\text{DetE}}^{-}(\hat{\boldsymbol{x}}_{t_{n-1}},t_{n-1},c),t_{0}), \\
\end{aligned}
\end{equation}
Then we perform the forward process and apply noise to them to obtain fake sample $\boldsymbol{x}_{fake}$ and real sample $\boldsymbol{x}_{real}$ with Eq.~\ref{eq:forward}. We use a frozen teacher model as the feature extraction backbone $\Omega$, extracting intermediate features with a fixed stride for calculating the GAN loss and Feature Matching loss $\mathcal{L}_{FM}$. During training, we iteratively update the parameters of $F_{\text{DetE}}$ and the discriminator head $f_D$:
\vspace{-0.5em}
\begin{equation}
\begin{aligned}
&\mathcal{L}_{FM}=\mathbb{E}_{\boldsymbol{x},t_n} \left\|\Omega(\boldsymbol{x}_{fake}) - \Omega(\boldsymbol{x}_{real}) \right\|^2_2, \\
&\mathcal{L}_G = \mathbb{E}_{\boldsymbol{x},t_n}[1-f_D(\Omega(\boldsymbol{x}_{fake}))] + \mathcal{L}_{FM}, \\
&\mathcal{L}_D = \mathbb{E}_{\boldsymbol{x},t_n}[f_D(\Omega(\boldsymbol{x}_{fake}))] + \mathbb{E}_{\boldsymbol{x},t_n}[1 - f_D(\Omega(\boldsymbol{x}_{real}))].
\end{aligned}
\end{equation}
The integration of the GAN loss in combination with Feature Matching loss provides a robust framework for training the details expert denoiser $F_{\text{DetE}}$, stabilizing its learning process and improving the quality of detail synthesis.
\section{Experiments}
\vspace{-0.3em}
\subsection{Experimental Setup}
\vspace{-0.3em}
\noindent\textbf{Backbones and Baselines}\quad We utilize HunyuanVideo~\cite{kong2024hunyuanvideo} and CogVideoX~\cite{yang2024cogvideox} as the base models for distillation. The HunyuanVideo has 13 billion parameters, and CogVideoX has 2 billion parameters. Since most prior distillation methods for diffusion models have not been applied to video synthesis, we follow the official implementations of LCM~\cite{luo2023latent} and PCM~\cite{wang2024phased} to implement these two methods on the selected base models as baselines for comparison.

\noindent\textbf{Implementation Details}\quad For HunyuanVideo, we selected trajectories with 50 Euler steps and used the default sampling parameters from diffusers. The distillation was conducted on 129-frame video sequences at a resolution of $1280\times720$ with a batch size of 6. For the semantic expert denoiser, we performed $1000$ iterations of distillation with a learning rate of $1e-6$, while for the details expert denoiser, we trained for $1000$ iterations with a learning rate of $5e-6$. For CogVideoX-2B, we selected trajectories with 50 DDIM steps. The distillation was conducted on 29-frame video sequences at a resolution of $720\times 480$ with a batch size of 4. In the first-stage fine-tuning, we distilled for approximately 1000 steps, while in the second-stage fine-tuning, we distilled for around 500 steps, both with a learning rate of $1e-6$. All experiments were conducted on 24 NVIDIA A100 80GB GPUs.

\noindent\textbf{Evaluation Metrics\quad}For video quality evaluation, we use VBench~\cite{huang2024vbench} as our assessment metric. VBench is a comprehensive benchmark suite for video generative models, designed to align closely with human perception and offer valuable insights from multiple perspectives.
Additionally, we conducted a user study to help evaluate the visual quality of the generated videos.

\vspace{-0.3em}
\subsection{Main Results}
\vspace{-0.3em}
\noindent\textbf{Quantitative Comparison}\quad Table~\ref{quant} presents the quantitative comparison of our method with LCM and PCM on HunyuanVideo and CogVideoX. We generate videos using the prompts provided by VBench to evaluate their performance in terms of semantic alignment and visual quality. It can be observed that on HunyuanVideo, our method achieves a VBench score comparable to the baseline with 4-step sampling, significantly outperforming LCM and PCM. In terms of efficiency, our method incurs a nearly identical latency cost per inference step compared to LCM and PCM.
\begin{figure*}[t]
\centering
\includegraphics[width=0.8\linewidth]{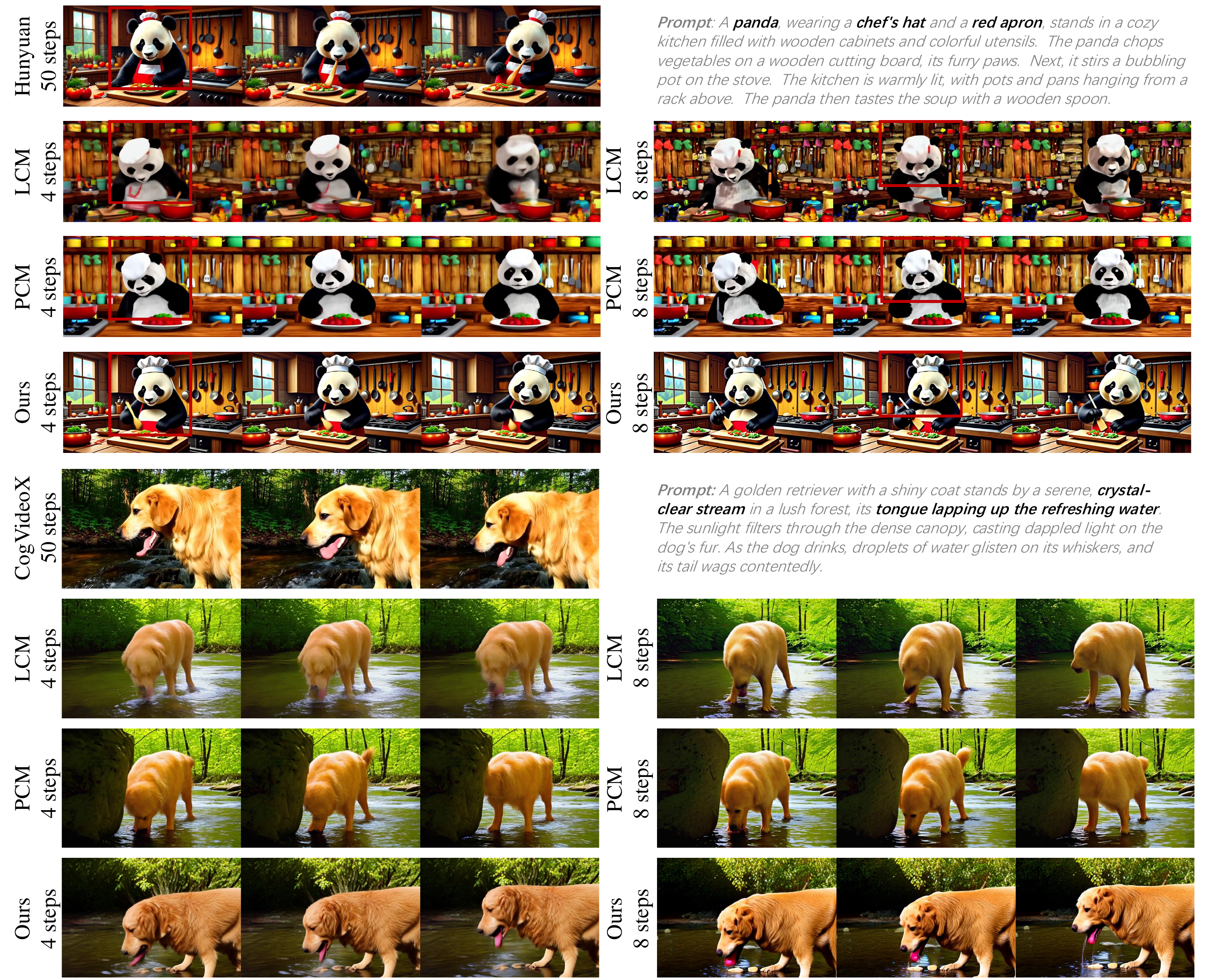}
\vspace{-0.3em}
\caption{Visual quality comparison of different methods. Differences are highlighted in boxes.}
\vspace{-1.75em}
\label{fig:quality}
\end{figure*}
\begin{table}[h]
    \centering
    \vspace{-1.em}
    \caption{Comparison of efficiency and visual quality of different methods. The latency of HunyuanVideo was measured on two A100 GPUs, and that of CogVideoX on a single A100 GPU.}
    \vspace{-0.75em}
    \begin{tabularx}{\linewidth}{c|c|c|X X X}
        \toprule
        \multirow{2}{*}{{Method}} & \multirow{2}{*}{{Step}} & \multirow{2}{*}{{Lat.(Sec.)}} & \multicolumn{3}{c}{{VBench}} \\ \cline{4-6}
        & & & \scriptsize{Total} & \scriptsize{Quality} & \scriptsize{Semantic} \\ \hline
        \rowcolor{lightgray}
        Hunyuan & 50 & 1504.5 & 83.87 & 85.00 & 79.34 \\ 
        LCM & 4 & 120.68 & 80.33 & 80.83 & 78.32 \\ 
        PCM & 4 & 120.89 & 80.93 & 81.94 & 76.90 \\ 
        Ours & 4 & 121.52 & \textbf{83.83} & \textbf{85.12} & \textbf{78.67} \\ 
        \hline
        LCM & 8 & 242.80 & 81.49 & 82.35 & 78.03 \\ 
        PCM & 8 & 242.96 & 81.63 & 82.78 & 77.00 \\ 
        Ours & 8 & 244.72 & \textbf{83.86} & \textbf{85.00} & \textbf{79.32} \\ 
        \hline
        \hline
        \rowcolor{lightgray}
        CogVideoX & 50 & 76.50 & 80.59& 81.93&75.23 \\ 
        LCM & 4 & 3.22& 78.88& 80.07&74.12 \\ 
        PCM & 4 & 3.23&79.09& 80.33&74.14 \\ 
        Ours & 4 &3.31 &\textbf{79.99}&\textbf{81.35} &\textbf{74.56} \\ 
        \hline
        LCM & 8 & 6.42& 79.34& 80.64&74.21 \\ 
        PCM & 8 & 6.42 & 79.70&80.98 &74.60 \\ 
        Ours & 8 & 6.58 & \textbf{80.26} & \textbf{81.57}&\textbf{75.03} \\ 
        \bottomrule
    \end{tabularx}
    \label{quant}
    \vspace{-0.8em}
\end{table}

\noindent\textbf{Qualitative Comparison}\quad Fig.~\ref{fig:quality} presents a comparison of the videos generated by our method and those produced by the original model, LCM and PCM. The results demonstrate that our method maintains high semantic and detail quality in synthesized videos while reducing the number of inference steps. Additional qualitative results are provided in the supplementary material for further reference.

\noindent\textbf{User Study}\quad To further evaluate the effectiveness of our method, we conduct a human evaluation to assess the perceived visual quality of the generated videos. Specifically, we randomly select 30 videos for each model. During the evaluation, each rater is presented with a text prompt along with two videos generated by different distillation methods, displayed in a randomized order to eliminate bias. Following the protocol of human preference evaluation in HunyuanVideo~\cite{kong2024hunyuanvideo}, the professional raters are asked to choose the video they perceive to have superior \textit{text alignment}, \textit{motion quality}, and \textit{visual quality}. Each sample is evaluated by fifty independent raters, and the aggregated voting results are summarized in Table~\ref{user}. As one can see, compared to other distillation methods, the raters significantly prefer the videos generated by our method.

\begin{table}[h]
    \vspace{-0.75em}
    \caption{User preference study. The numbers represent the percentage of raters who favor the videos synthesized by our method.}
    \vspace{-0.75em}
    \centering
    \begin{tabular}{l|c c}
        \toprule
        Method comparison & HunyuanVideo & CogVideoX \\
        \hline
        Ours vs.  LCM & 82.67\% & 75.33\% \\
        \hline
        Ours vs.  PCM & 77.33\%  & 72.67\% \\
        \bottomrule
    \end{tabular}
    \label{user}
    \vspace{-0.5em}
\end{table}
\vspace{-0.75em}
\subsection{Ablation Study}
To thoroughly evaluate both the effectiveness of our method, we conduct extensive ablation studies based on HunyuanVideo, as shown in Table~\ref{abatable}. All experiments were conducted on 29-frame videos with a resolution of $1280\times 720$. Inference is performed with 4 sampling timesteps.
\begin{table}[h]
    \centering
    \vspace{-0.75em}
    \caption{Impact of different components of our method.}
    \vspace{-0.5em}
    \begin{tabularx}{\linewidth}{c|X X X X|X X X}
        \toprule
        \multirow{2}{*}{} & \multicolumn{4}{c|}{{Variants}} & \multicolumn{3}{c}{{VBench}} \\ 
        & OD & PE & TC & GF & \scriptsize{Total} & \scriptsize{Quality} & \scriptsize{Semantic} \\ \hline

        (1) & & & & &  80.30 & 80.74 & 78.36 \\
        (2) &\checkmark & & & & 83.08 & 84.20 & 78.59 \\
        (3) &\checkmark &\checkmark & & & 83.03 & 84.16 & 78.53 \\
        (4) &\checkmark &\checkmark &\checkmark & & 83.42 & 84.63 & \textbf{78.63} \\
        (5) &\checkmark &\checkmark & &\checkmark & 83.71 & 84.99 & 78.59 \\
        (6) &\checkmark &\checkmark &\checkmark &\checkmark & \textbf{83.80} & \textbf{85.10} & 78.62 \\   
        \bottomrule
    \end{tabularx}
    \label{abatable}
    \vspace{-1.em}
\end{table}

\begin{figure}[t]
    \centering
    \includegraphics[width=0.95\columnwidth]{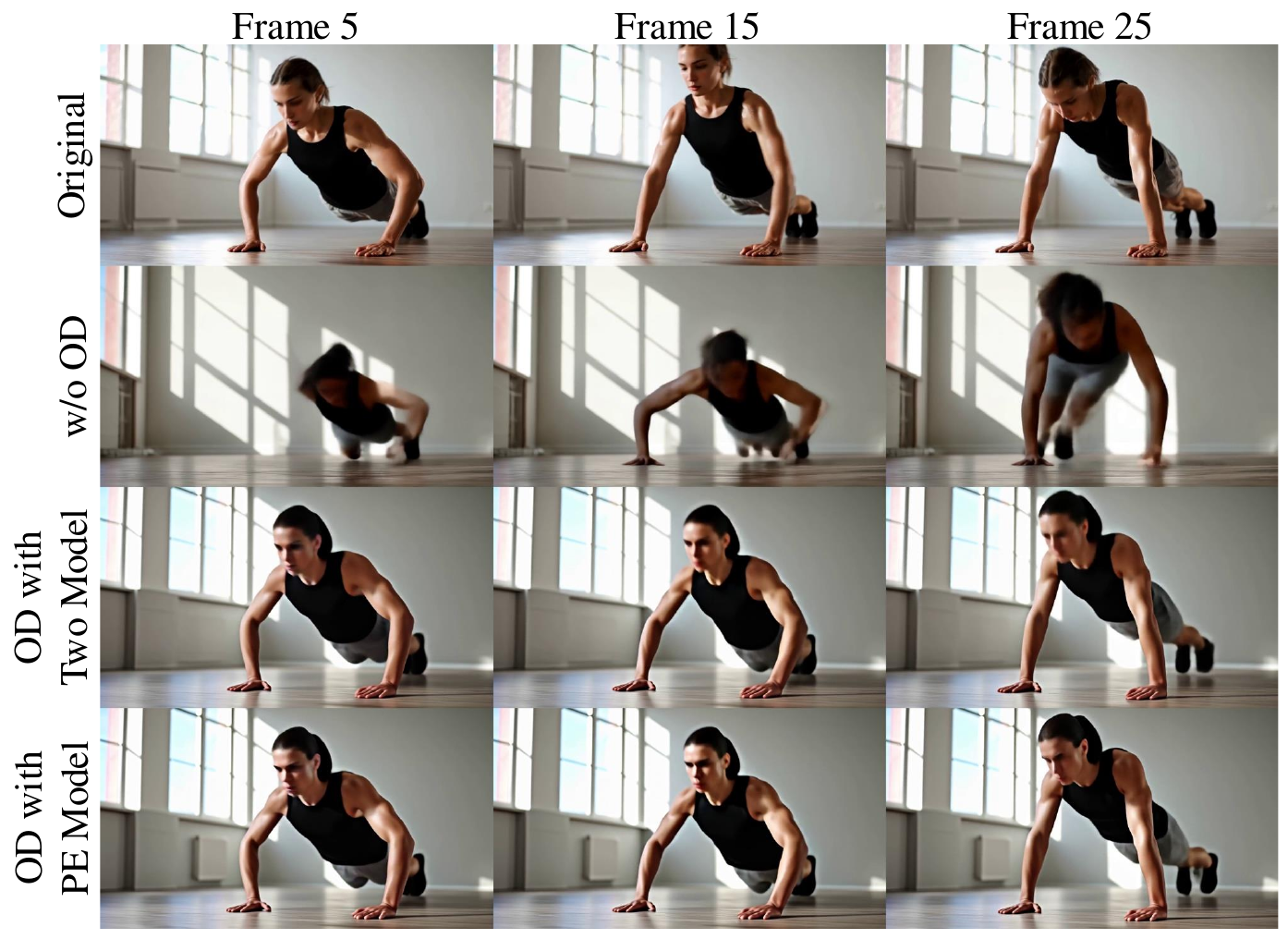}
    \vspace{-0.75em}
    \caption{Impact of optimization decoupling and parameter-efficient distillation.}
    \label{fig:decouple}
    \vspace{-0.8em}
\end{figure}
\noindent\textbf{Effect of Optimization Decoupling (OD)}\quad
Through Experiments (1) and (2), decoupling the optimization of semantic and detail modeling significantly improves the semantic and quality scores. As shown in Fig.~\ref{fig:decouple}, the optimized decoupled model synthesizes videos with better semantic and detail quality, where the motion of characters and facial details appear more natural.

\noindent\textbf{Effect of Parameter-Efficient dual-expert distillation (PE)}\quad Through Experiments (2) and (3), we observe that compared to simply decoupling the optimization into two separate model training processes, the parameter-efficient Dual-Expert distillation significantly reduces both the parameters and memory requirements, with minimal computational overhead, while preserving visual quality. The last two rows of Fig.~\ref{fig:decouple} also demonstrate that our parameter-efficient Dual-Expert method does not result in a significant degradation in visual quality.

\begin{figure}[t]
    \centering
    \includegraphics[width=0.92\columnwidth]{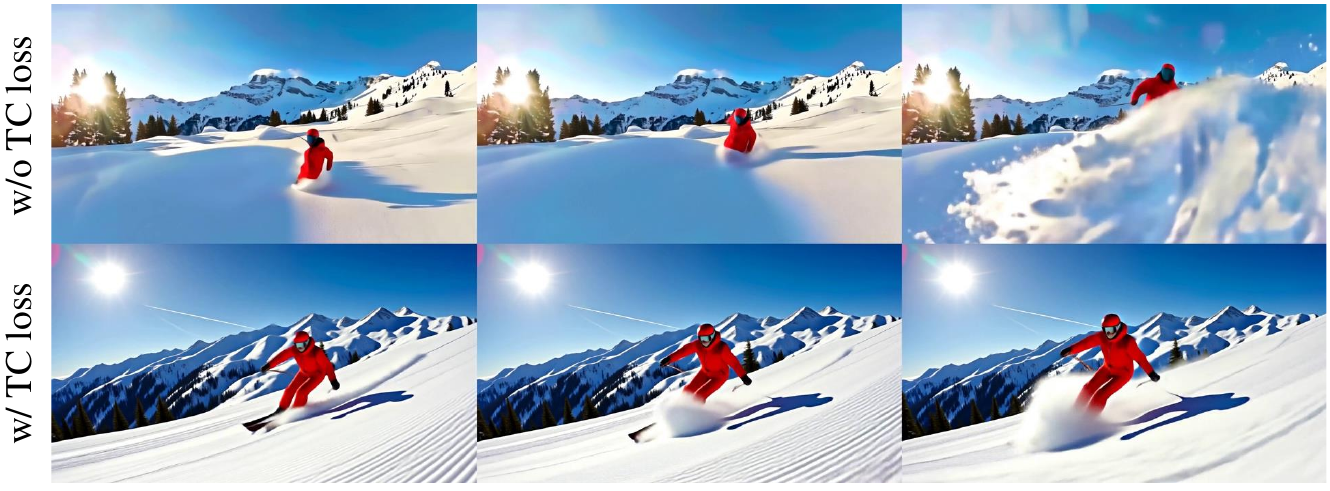}
    \vspace{-0.75em}
    \caption{Impact of temporal coherence loss.}
    \vspace{-0.75em}
    \label{fig:tcl}
\end{figure}
\noindent\textbf{Effect of Temporal Coherence Loss (TC)}\quad By comparing Experiments (3) and (4), or (5) and (6), we observe that the introduction of Temporal Coherence loss improves the quality scores of the synthesized videos. As shown in Fig.~\ref{fig:tcl}, the introduction of the TC loss enables more natural motion in the video and enhances its consistency.

\begin{figure}[t]
    \centering
    \includegraphics[width=0.92\columnwidth]{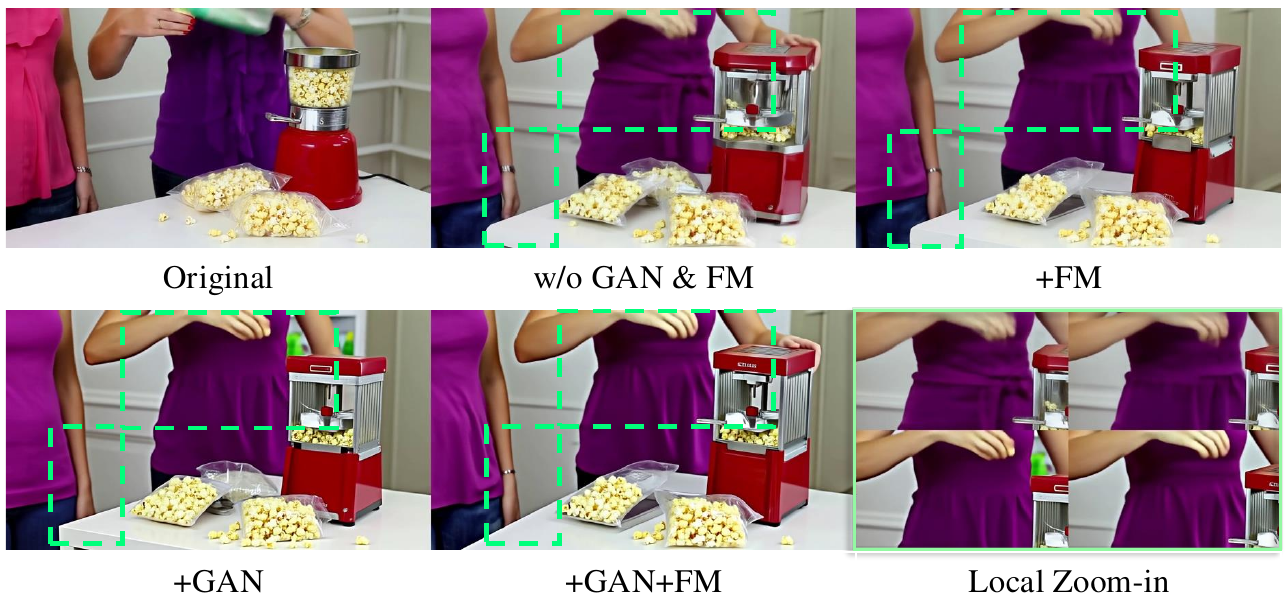}
    \vspace{-1em}
    \caption{Impact of the GAN loss and Feature Matching term.}
    \vspace{-1.5em}
    \label{fig:gan}
\end{figure}
\noindent\textbf{Effect of GAN and Feature Matching Loss (GF)}\quad By comparing Experiments (3) and (5), or (4) and (6), we observe that the introduction of GAN Loss improves the quality scores of the synthesized videos. As shown in Fig.~\ref{fig:gan}, the introduction of the GAN loss and Feature Matching term enhances the realism of the details in the synthesized video.

\begin{figure}[t]
    \centering
    \includegraphics[width=\columnwidth]{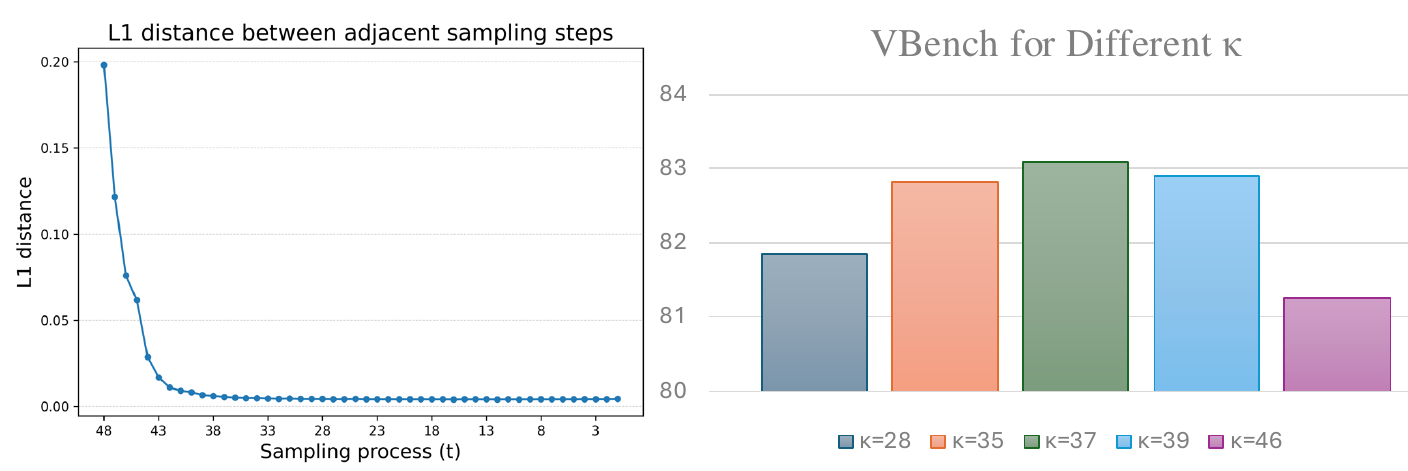}
    \vspace{-2em}
    \caption{Impact of different $\kappa$.}
    \vspace{-2em}
    \label{fig:kappa}
\end{figure}
\noindent\textbf{Selection of $\kappa$} In this paper, we determine the value of $\kappa$ based on the inference process. Fig.~\ref{fig:kappa} (left) illustrates the L1 distance between adjacent time-step sampling results in HunyuanVideo during the sampling process. It can be observed that from approximately step 37 onward, the L1 Distance decreases to a very small value. We interpret this as the point where the semantic content and layout are established, and the remaining steps focus on synthesizing high-frequency details. Therefore, we set $\kappa=37$ as the default value.
To evaluate the impact of different $\kappa$ values, we experimented with $\kappa=28,35,37,39, 46$. 
As shown in Fig.~\ref{fig:kappa} (right), the results indicate that as $\kappa$ deviates from the transition point between semantic synthesis and detail synthesis, the video quality gradually deteriorates. It further validates the effectiveness of our optimization decoupling strategy.

\vspace{-0.5em}
\section{Conclusion and Discussion}
In this paper, we identify a key optimization conflict in consistency distillation for video synthesis: there exists a significant discrepancy in the optimization gradients and loss contributions across different timesteps. Distilling the entire ODE trajectory into a single student model fails to balance these aspects, leading to degraded motion consistency and coarse synthesis quality.
To address this issue, we propose a parameter-efficient Dual-Expert distillation framework that decouples semantic learning from fine-detail refinement. Additionally, we introduce a Temporal Coherence loss to enhance motion consistency for the semantic expert and apply GAN and Feature Matching loss to improve synthesis quality for the detail expert. Our method significantly reduces sampling steps while achieving state-of-the-art visual quality, demonstrating the effectiveness of expert specialization in video diffusion model distillation.

\noindent\textbf{Limitation}\quad Although our method achieves favorable results with 4 steps inference, it still struggles to produce satisfactory outcomes with fewer steps (e.g., 2) due to limited training data and iterations. We will further explore high-quality synthesis with fewer steps in future work.
\vspace{-0.5em}
\section{Acknowledgement}
This study is supported by the Ministry of Education, Singapore, under its MOE AcRF Tier 2 (MOE-T2EP20221-0012, MOE-T2EP20223-0002), and under the RIE2020 Industry Alignment Fund – Industry Collaboration Projects (IAF-ICP) Funding Initiative, as well as cash and in-kind contribution from the industry partner(s).

{
    \small
    \bibliographystyle{ieeenat_fullname}
    \bibliography{main}

\begin{thebibliography}{76}
\providecommand{\natexlab}[1]{#1}
\providecommand{\url}[1]{\texttt{#1}}
\expandafter\ifx\csname urlstyle\endcsname\relax
  \providecommand{\doi}[1]{doi: #1}\else
  \providecommand{\doi}{doi: \begingroup \urlstyle{rm}\Url}\fi

\bibitem[Berthelot et~al.(2023)Berthelot, Autef, Lin, Yap, Zhai, Hu, Zheng, Talbott, and Gu]{berthelot2023tract}
David Berthelot, Arnaud Autef, Jierui Lin, Dian~Ang Yap, Shuangfei Zhai, Siyuan Hu, Daniel Zheng, Walter Talbott, and Eric Gu.
\newblock Tract: Denoising diffusion models with transitive closure time-distillation.
\newblock \emph{arXiv preprint arXiv:2303.04248}, 2023.

\bibitem[Blattmann et~al.(2023)Blattmann, Rombach, Ling, Dockhorn, Kim, Fidler, and Kreis]{blattmann2023align}
Andreas Blattmann, Robin Rombach, Huan Ling, Tim Dockhorn, Seung~Wook Kim, Sanja Fidler, and Karsten Kreis.
\newblock Align your latents: High-resolution video synthesis with latent diffusion models.
\newblock In \emph{Proceedings of the IEEE/CVF Conference on Computer Vision and Pattern Recognition}, pages 22563--22575, 2023.

\bibitem[Chen et~al.(2024)Chen, Bandyopadhyay, Zou, and Song]{chen2024nitrofusion}
Dar-Yen Chen, Hmrishav Bandyopadhyay, Kai Zou, and Yi-Zhe Song.
\newblock Nitrofusion: High-fidelity single-step diffusion through dynamic adversarial training.
\newblock \emph{arXiv preprint arXiv:2412.02030}, 2024.

\bibitem[Goodfellow et~al.(2014)Goodfellow, Pouget-Abadie, Mirza, Xu, Warde-Farley, Ozair, Courville, and Bengio]{goodfellow2014generative}
Ian Goodfellow, Jean Pouget-Abadie, Mehdi Mirza, Bing Xu, David Warde-Farley, Sherjil Ozair, Aaron Courville, and Yoshua Bengio.
\newblock Generative adversarial nets.
\newblock \emph{Advances in neural information processing systems}, 27, 2014.

\bibitem[Gu et~al.(2023)Gu, Zhai, Zhang, Liu, and Susskind]{gu2023boot}
Jiatao Gu, Shuangfei Zhai, Yizhe Zhang, Lingjie Liu, and Joshua~M Susskind.
\newblock Boot: Data-free distillation of denoising diffusion models with bootstrapping.
\newblock In \emph{ICML 2023 Workshop on Structured Probabilistic Inference $\{$$\backslash$\&$\}$ Generative Modeling}, 2023.

\bibitem[Guo et~al.(2023)Guo, Yang, Rao, Liang, Wang, Qiao, Agrawala, Lin, and Dai]{guo2023animatediff}
Yuwei Guo, Ceyuan Yang, Anyi Rao, Zhengyang Liang, Yaohui Wang, Yu Qiao, Maneesh Agrawala, Dahua Lin, and Bo Dai.
\newblock Animatediff: Animate your personalized text-to-image diffusion models without specific tuning.
\newblock \emph{arXiv preprint arXiv:2307.04725}, 2023.

\bibitem[HaCohen et~al.(2024)HaCohen, Chiprut, Brazowski, Shalem, Moshe, Richardson, Levin, Shiran, Zabari, Gordon, Panet, Weissbuch, Kulikov, Bitterman, Melumian, and Bibi]{HaCohen2024LTXVideo}
Yoav HaCohen, Nisan Chiprut, Benny Brazowski, Daniel Shalem, Dudu Moshe, Eitan Richardson, Eran Levin, Guy Shiran, Nir Zabari, Ori Gordon, Poriya Panet, Sapir Weissbuch, Victor Kulikov, Yaki Bitterman, Zeev Melumian, and Ofir Bibi.
\newblock Ltx-video: Realtime video latent diffusion.
\newblock \emph{arXiv preprint arXiv:2501.00103}, 2024.

\bibitem[Ho et~al.(2020)Ho, Jain, and Abbeel]{ho2020denoising}
Jonathan Ho, Ajay Jain, and Pieter Abbeel.
\newblock Denoising diffusion probabilistic models.
\newblock \emph{Advances in neural information processing systems}, 33:\penalty0 6840--6851, 2020.

\bibitem[Ho et~al.(2022)Ho, Salimans, Gritsenko, Chan, Norouzi, and Fleet]{ho2022video}
Jonathan Ho, Tim Salimans, Alexey Gritsenko, William Chan, Mohammad Norouzi, and David~J Fleet.
\newblock Video diffusion models.
\newblock \emph{Advances in Neural Information Processing Systems}, 35:\penalty0 8633--8646, 2022.

\bibitem[Hong et~al.(2022)Hong, Ding, Zheng, Liu, and Tang]{hong2022cogvideo}
Wenyi Hong, Ming Ding, Wendi Zheng, Xinghan Liu, and Jie Tang.
\newblock Cogvideo: Large-scale pretraining for text-to-video generation via transformers.
\newblock \emph{arXiv preprint arXiv:2205.15868}, 2022.

\bibitem[Hu et~al.(2021)Hu, Shen, Wallis, Allen-Zhu, Li, Wang, Wang, and Chen]{hu2021lora}
Edward~J Hu, Yelong Shen, Phillip Wallis, Zeyuan Allen-Zhu, Yuanzhi Li, Shean Wang, Lu Wang, and Weizhu Chen.
\newblock Lora: Low-rank adaptation of large language models.
\newblock \emph{arXiv preprint arXiv:2106.09685}, 2021.

\bibitem[Hu(2024)]{hu2024animate}
Li Hu.
\newblock Animate anyone: Consistent and controllable image-to-video synthesis for character animation.
\newblock In \emph{Proceedings of the IEEE/CVF Conference on Computer Vision and Pattern Recognition}, pages 8153--8163, 2024.

\bibitem[Huang et~al.(2025)Huang, Zhang, Zeng, Zhang, Li, Zuo, and Lau]{huang2025dreamphysics}
Tianyu Huang, Haoze Zhang, Yihan Zeng, Zhilu Zhang, Hui Li, Wangmeng Zuo, and Rynson~WH Lau.
\newblock Dreamphysics: Learning physics-based 3d dynamics with video diffusion priors.
\newblock In \emph{Proceedings of the AAAI Conference on Artificial Intelligence}, pages 3733--3741, 2025.

\bibitem[Huang et~al.(2024)Huang, He, Yu, Zhang, Si, Jiang, Zhang, Wu, Jin, Chanpaisit, et~al.]{huang2024vbench}
Ziqi Huang, Yinan He, Jiashuo Yu, Fan Zhang, Chenyang Si, Yuming Jiang, Yuanhan Zhang, Tianxing Wu, Qingyang Jin, Nattapol Chanpaisit, et~al.
\newblock Vbench: Comprehensive benchmark suite for video generative models.
\newblock In \emph{Proceedings of the IEEE/CVF Conference on Computer Vision and Pattern Recognition}, pages 21807--21818, 2024.

\bibitem[Jin et~al.(2024)Jin, Sun, Li, Xu, Jiang, Zhuang, Huang, Song, Mu, and Lin]{jin2024pyramidal}
Yang Jin, Zhicheng Sun, Ningyuan Li, Kun Xu, Hao Jiang, Nan Zhuang, Quzhe Huang, Yang Song, Yadong Mu, and Zhouchen Lin.
\newblock Pyramidal flow matching for efficient video generative modeling.
\newblock \emph{arXiv preprint arXiv:2410.05954}, 2024.

\bibitem[Kang et~al.(2024)Kang, Zhang, Barnes, Paris, Kwak, Park, Shechtman, Zhu, and Park]{kang2024distilling}
Minguk Kang, Richard Zhang, Connelly Barnes, Sylvain Paris, Suha Kwak, Jaesik Park, Eli Shechtman, Jun-Yan Zhu, and Taesung Park.
\newblock Distilling diffusion models into conditional gans.
\newblock In \emph{European Conference on Computer Vision}, pages 428--447. Springer, 2024.

\bibitem[Karras et~al.(2022)Karras, Aittala, Aila, and Laine]{karras2022elucidating}
Tero Karras, Miika Aittala, Timo Aila, and Samuli Laine.
\newblock Elucidating the design space of diffusion-based generative models.
\newblock 35:\penalty0 26565--26577, 2022.

\bibitem[Kim et~al.(2023)Kim, Lai, Liao, Murata, Takida, Uesaka, He, Mitsufuji, and Ermon]{kim2023consistency}
Dongjun Kim, Chieh-Hsin Lai, Wei-Hsiang Liao, Naoki Murata, Yuhta Takida, Toshimitsu Uesaka, Yutong He, Yuki Mitsufuji, and Stefano Ermon.
\newblock Consistency trajectory models: Learning probability flow ode trajectory of diffusion.
\newblock \emph{arXiv preprint arXiv:2310.02279}, 2023.

\bibitem[Kohler et~al.(2024)Kohler, Pumarola, Sch{\"o}nfeld, Sanakoyeu, Sumbaly, Vajda, and Thabet]{kohler2024imagine}
Jonas Kohler, Albert Pumarola, Edgar Sch{\"o}nfeld, Artsiom Sanakoyeu, Roshan Sumbaly, Peter Vajda, and Ali Thabet.
\newblock Imagine flash: Accelerating emu diffusion models with backward distillation.
\newblock \emph{arXiv preprint arXiv:2405.05224}, 2024.

\bibitem[Kong et~al.(2024)Kong, Tian, Zhang, Min, Dai, Zhou, Xiong, Li, Wu, Zhang, et~al.]{kong2024hunyuanvideo}
Weijie Kong, Qi Tian, Zijian Zhang, Rox Min, Zuozhuo Dai, Jin Zhou, Jiangfeng Xiong, Xin Li, Bo Wu, Jianwei Zhang, et~al.
\newblock Hunyuanvideo: A systematic framework for large video generative models.
\newblock \emph{arXiv preprint arXiv:2412.03603}, 2024.

\bibitem[Kuaishou(2024)]{kuaishou}
Kuaishou.
\newblock Kling, 2024.

\bibitem[Lab and etc.(2024)]{pku_yuan_lab_and_tuzhan_ai_etc_2024_10948109}
PKU-Yuan Lab and Tuzhan~AI etc.
\newblock Open-sora-plan, 2024.

\bibitem[Lee et~al.(2024)Lee, Xu, Geffner, Fanti, Kreis, Vahdat, and Nie]{lee2024truncated}
Sangyun Lee, Yilun Xu, Tomas Geffner, Giulia Fanti, Karsten Kreis, Arash Vahdat, and Weili Nie.
\newblock Truncated consistency models.
\newblock \emph{arXiv preprint arXiv:2410.14895}, 2024.

\bibitem[Lin and Yang(2024)]{lin2024animatediff}
Shanchuan Lin and Xiao Yang.
\newblock Animatediff-lightning: Cross-model diffusion distillation.
\newblock \emph{arXiv preprint arXiv:2403.12706}, 2024.

\bibitem[Lin et~al.(2024)Lin, Wang, and Yang]{lin2024sdxl}
Shanchuan Lin, Anran Wang, and Xiao Yang.
\newblock Sdxl-lightning: Progressive adversarial diffusion distillation.
\newblock \emph{arXiv preprint arXiv:2402.13929}, 2024.

\bibitem[Lin et~al.(2025)Lin, Xia, Ren, Yang, Xiao, and Jiang]{lin2025diffusion}
Shanchuan Lin, Xin Xia, Yuxi Ren, Ceyuan Yang, Xuefeng Xiao, and Lu Jiang.
\newblock Diffusion adversarial post-training for one-step video generation.
\newblock \emph{arXiv preprint arXiv:2501.08316}, 2025.

\bibitem[Liu et~al.(2024)Liu, Xie, Deng, Chen, Tang, Fu, Zha, and Lu]{liu2024scott}
Hongjian Liu, Qingsong Xie, Zhijie Deng, Chen Chen, Shixiang Tang, Fueyang Fu, Zheng-jun Zha, and Haonan Lu.
\newblock Scott: Accelerating diffusion models with stochastic consistency distillation.
\newblock \emph{arXiv preprint arXiv:2403.01505}, 2024.

\bibitem[Liu et~al.(2022{\natexlab{a}})Liu, Ren, Lin, and Zhao]{liu2022pseudo}
Luping Liu, Yi Ren, Zhijie Lin, and Zhou Zhao.
\newblock Pseudo numerical methods for diffusion models on manifolds.
\newblock \emph{arXiv preprint arXiv:2202.09778}, 2022{\natexlab{a}}.

\bibitem[Liu et~al.(2022{\natexlab{b}})Liu, Gong, and Liu]{liu2022flow}
Xingchao Liu, Chengyue Gong, and Qiang Liu.
\newblock Flow straight and fast: Learning to generate and transfer data with rectified flow.
\newblock \emph{arXiv preprint arXiv:2209.03003}, 2022{\natexlab{b}}.

\bibitem[Liu et~al.(2023)Liu, Zhang, Ma, Peng, et~al.]{liu2023instaflow}
Xingchao Liu, Xiwen Zhang, Jianzhu Ma, Jian Peng, et~al.
\newblock Instaflow: One step is enough for high-quality diffusion-based text-to-image generation.
\newblock In \emph{The Twelfth International Conference on Learning Representations}, 2023.

\bibitem[Lu and Song(2024)]{lu2024simplifying}
Cheng Lu and Yang Song.
\newblock Simplifying, stabilizing and scaling continuous-time consistency models.
\newblock \emph{arXiv preprint arXiv:2410.11081}, 2024.

\bibitem[Lu et~al.(2022{\natexlab{a}})Lu, Zhou, Bao, Chen, Li, and Zhu]{lu2022dpm}
Cheng Lu, Yuhao Zhou, Fan Bao, Jianfei Chen, Chongxuan Li, and Jun Zhu.
\newblock Dpm-solver: A fast ode solver for diffusion probabilistic model sampling in around 10 steps.
\newblock \emph{Advances in Neural Information Processing Systems}, 35:\penalty0 5775--5787, 2022{\natexlab{a}}.

\bibitem[Lu et~al.(2022{\natexlab{b}})Lu, Zhou, Bao, Chen, Li, and Zhu]{lu2022dpmpp}
Cheng Lu, Yuhao Zhou, Fan Bao, Jianfei Chen, Chongxuan Li, and Jun Zhu.
\newblock Dpm-solver++: Fast solver for guided sampling of diffusion probabilistic models.
\newblock \emph{arXiv preprint arXiv:2211.01095}, 2022{\natexlab{b}}.

\bibitem[Luhman and Luhman(2021)]{luhman2021knowledge}
Eric Luhman and Troy Luhman.
\newblock Knowledge distillation in iterative generative models for improved sampling speed.
\newblock \emph{arXiv preprint arXiv:2101.02388}, 2021.

\bibitem[Luo et~al.(2023{\natexlab{a}})Luo, Tan, Huang, Li, and Zhao]{luo2023latent}
Simian Luo, Yiqin Tan, Longbo Huang, Jian Li, and Hang Zhao.
\newblock Latent consistency models: Synthesizing high-resolution images with few-step inference.
\newblock \emph{arXiv preprint arXiv:2310.04378}, 2023{\natexlab{a}}.

\bibitem[Luo et~al.(2023{\natexlab{b}})Luo, Tan, Patil, Gu, von Platen, Passos, Huang, Li, and Zhao]{luo2023lcm}
Simian Luo, Yiqin Tan, Suraj Patil, Daniel Gu, Patrick von Platen, Apolin{\'a}rio Passos, Longbo Huang, Jian Li, and Hang Zhao.
\newblock Lcm-lora: A universal stable-diffusion acceleration module.
\newblock \emph{arXiv preprint arXiv:2311.05556}, 2023{\natexlab{b}}.

\bibitem[Luo(2023)]{luo2023comprehensive}
Weijian Luo.
\newblock A comprehensive survey on knowledge distillation of diffusion models.
\newblock \emph{arXiv preprint arXiv:2304.04262}, 2023.

\bibitem[Luo et~al.(2023{\natexlab{c}})Luo, Hu, Zhang, Sun, Li, and Zhang]{luo2023diff}
Weijian Luo, Tianyang Hu, Shifeng Zhang, Jiacheng Sun, Zhenguo Li, and Zhihua Zhang.
\newblock Diff-instruct: A universal approach for transferring knowledge from pre-trained diffusion models.
\newblock \emph{Advances in Neural Information Processing Systems}, 36:\penalty0 76525--76546, 2023{\natexlab{c}}.

\bibitem[Luo et~al.(2024{\natexlab{a}})Luo, Huang, Geng, Kolter, and Qi]{luo2024one}
Weijian Luo, Zemin Huang, Zhengyang Geng, J~Zico Kolter, and Guo-jun Qi.
\newblock One-step diffusion distillation through score implicit matching.
\newblock \emph{arXiv preprint arXiv:2410.16794}, 2024{\natexlab{a}}.

\bibitem[Luo et~al.(2024{\natexlab{b}})Luo, Chen, Qu, Hu, and Tang]{luo2024you}
Yihong Luo, Xiaolong Chen, Xinghua Qu, Tianyang Hu, and Jing Tang.
\newblock You only sample once: Taming one-step text-to-image synthesis by self-cooperative diffusion gans.
\newblock \emph{arXiv preprint arXiv:2403.12931}, 2024{\natexlab{b}}.

\bibitem[Lv et~al.(2024)Lv, Si, Song, Yang, Qiao, Liu, and Wong]{lv2024fastercache}
Zhengyao Lv, Chenyang Si, Junhao Song, Zhenyu Yang, Yu Qiao, Ziwei Liu, and Kwan-Yee~K Wong.
\newblock Fastercache: Training-free video diffusion model acceleration with high quality.
\newblock \emph{arXiv preprint arXiv:2410.19355}, 2024.

\bibitem[Ma et~al.(2024)Ma, Wang, Jia, Chen, Liu, Li, Chen, and Qiao]{ma2024latte}
Xin Ma, Yaohui Wang, Gengyun Jia, Xinyuan Chen, Ziwei Liu, Yuan-Fang Li, Cunjian Chen, and Yu Qiao.
\newblock Latte: Latent diffusion transformer for video generation.
\newblock \emph{arXiv preprint arXiv:2401.03048}, 2024.

\bibitem[Meng et~al.(2023)Meng, Rombach, Gao, Kingma, Ermon, Ho, and Salimans]{meng2023distillation}
Chenlin Meng, Robin Rombach, Ruiqi Gao, Diederik Kingma, Stefano Ermon, Jonathan Ho, and Tim Salimans.
\newblock On distillation of guided diffusion models.
\newblock In \emph{Proceedings of the IEEE/CVF Conference on Computer Vision and Pattern Recognition}, pages 14297--14306, 2023.

\bibitem[OpenAI(2024)]{opensora}
OpenAI.
\newblock Sora, 2024.

\bibitem[Peebles and Xie(2023)]{peebles2023scalable}
William Peebles and Saining Xie.
\newblock Scalable diffusion models with transformers.
\newblock In \emph{Proceedings of the IEEE/CVF International Conference on Computer Vision}, pages 4195--4205, 2023.

\bibitem[Poole et~al.(2022)Poole, Jain, Barron, and Mildenhall]{poole2022dreamfusion}
Ben Poole, Ajay Jain, Jonathan~T Barron, and Ben Mildenhall.
\newblock Dreamfusion: Text-to-3d using 2d diffusion.
\newblock \emph{arXiv preprint arXiv:2209.14988}, 2022.

\bibitem[Ren et~al.(2024)Ren, Xia, Lu, Zhang, Wu, Xie, Wang, and Xiao]{ren2024hyper}
Yuxi Ren, Xin Xia, Yanzuo Lu, Jiacheng Zhang, Jie Wu, Pan Xie, Xing Wang, and Xuefeng Xiao.
\newblock Hyper-sd: Trajectory segmented consistency model for efficient image synthesis.
\newblock \emph{arXiv preprint arXiv:2404.13686}, 2024.

\bibitem[Rombach et~al.(2022)Rombach, Blattmann, Lorenz, Esser, and Ommer]{rombach2022high}
Robin Rombach, Andreas Blattmann, Dominik Lorenz, Patrick Esser, and Bj{\"o}rn Ommer.
\newblock High-resolution image synthesis with latent diffusion models.
\newblock In \emph{Proceedings of the IEEE/CVF conference on computer vision and pattern recognition}, pages 10684--10695, 2022.

\bibitem[Salimans and Ho(2022)]{salimans2022progressive}
Tim Salimans and Jonathan Ho.
\newblock Progressive distillation for fast sampling of diffusion models.
\newblock \emph{arXiv preprint arXiv:2202.00512}, 2022.

\bibitem[Sauer et~al.(2024{\natexlab{a}})Sauer, Boesel, Dockhorn, Blattmann, Esser, and Rombach]{sauer2024fast}
Axel Sauer, Frederic Boesel, Tim Dockhorn, Andreas Blattmann, Patrick Esser, and Robin Rombach.
\newblock Fast high-resolution image synthesis with latent adversarial diffusion distillation.
\newblock In \emph{SIGGRAPH Asia 2024 Conference Papers}, pages 1--11, 2024{\natexlab{a}}.

\bibitem[Sauer et~al.(2024{\natexlab{b}})Sauer, Lorenz, Blattmann, and Rombach]{sauer2024adversarial}
Axel Sauer, Dominik Lorenz, Andreas Blattmann, and Robin Rombach.
\newblock Adversarial diffusion distillation.
\newblock In \emph{European Conference on Computer Vision}, pages 87--103. Springer, 2024{\natexlab{b}}.

\bibitem[Song et~al.(2020)Song, Meng, and Ermon]{song2020denoising}
Jiaming Song, Chenlin Meng, and Stefano Ermon.
\newblock Denoising diffusion implicit models.
\newblock \emph{arXiv preprint arXiv:2010.02502}, 2020.

\bibitem[Song and Dhariwal(2023)]{song2023improved}
Yang Song and Prafulla Dhariwal.
\newblock Improved techniques for training consistency models.
\newblock \emph{arXiv preprint arXiv:2310.14189}, 2023.

\bibitem[Song et~al.(2023)Song, Dhariwal, Chen, and Sutskever]{song2023consistency}
Yang Song, Prafulla Dhariwal, Mark Chen, and Ilya Sutskever.
\newblock Consistency models.
\newblock \emph{arXiv preprint arXiv:2303.01469}, 2023.

\bibitem[Team(2024)]{genmo2024mochi}
Genmo Team.
\newblock Mochi 1.
\newblock \url{https://github.com/genmoai/models}, 2024.

\bibitem[Wan et~al.(2025)Wan, Wang, Ai, Wen, Mao, Xie, Chen, Yu, Zhao, Yang, et~al.]{wan2025wan}
Team Wan, Ang Wang, Baole Ai, Bin Wen, Chaojie Mao, Chen-Wei Xie, Di Chen, Feiwu Yu, Haiming Zhao, Jianxiao Yang, et~al.
\newblock Wan: Open and advanced large-scale video generative models.
\newblock \emph{arXiv preprint arXiv:2503.20314}, 2025.

\bibitem[Wang et~al.(2024{\natexlab{a}})Wang, Huang, Bergman, Shen, Gao, Lingelbach, Sun, Bian, Song, Liu, et~al.]{wang2024phased}
Fu-Yun Wang, Zhaoyang Huang, Alexander~William Bergman, Dazhong Shen, Peng Gao, Michael Lingelbach, Keqiang Sun, Weikang Bian, Guanglu Song, Yu Liu, et~al.
\newblock Phased consistency model.
\newblock \emph{arXiv preprint arXiv:2405.18407}, 2024{\natexlab{a}}.

\bibitem[Wang et~al.(2024{\natexlab{b}})Wang, Huang, Bian, Shi, Sun, Song, Liu, and Li]{wang2024animatelcm}
Fu-Yun Wang, Zhaoyang Huang, Weikang Bian, Xiaoyu Shi, Keqiang Sun, Guanglu Song, Yu Liu, and Hongsheng Li.
\newblock Animatelcm: Computation-efficient personalized style video generation without personalized video data.
\newblock In \emph{SIGGRAPH Asia 2024 Technical Communications}, pages 1--5. 2024{\natexlab{b}}.

\bibitem[Wang et~al.(2023)Wang, Yuan, Chen, Zhang, Wang, and Zhang]{wang2023modelscope}
Jiuniu Wang, Hangjie Yuan, Dayou Chen, Yingya Zhang, Xiang Wang, and Shiwei Zhang.
\newblock Modelscope text-to-video technical report.
\newblock \emph{arXiv preprint arXiv:2308.06571}, 2023.

\bibitem[Xi et~al.(2025)Xi, Yang, Zhao, Xu, Li, Li, Lin, Cai, Zhang, Li, et~al.]{xi2025sparse}
Haocheng Xi, Shuo Yang, Yilong Zhao, Chenfeng Xu, Muyang Li, Xiuyu Li, Yujun Lin, Han Cai, Jintao Zhang, Dacheng Li, et~al.
\newblock Sparse videogen: Accelerating video diffusion transformers with spatial-temporal sparsity.
\newblock \emph{arXiv preprint arXiv:2502.01776}, 2025.

\bibitem[Xiao et~al.(2021)Xiao, Kreis, and Vahdat]{xiao2021tackling}
Zhisheng Xiao, Karsten Kreis, and Arash Vahdat.
\newblock Tackling the generative learning trilemma with denoising diffusion gans.
\newblock \emph{arXiv preprint arXiv:2112.07804}, 2021.

\bibitem[Xu et~al.(2024)Xu, Zhao, Xiao, and Hou]{xu2024ufogen}
Yanwu Xu, Yang Zhao, Zhisheng Xiao, and Tingbo Hou.
\newblock Ufogen: You forward once large scale text-to-image generation via diffusion gans.
\newblock In \emph{Proceedings of the IEEE/CVF Conference on Computer Vision and Pattern Recognition}, pages 8196--8206, 2024.

\bibitem[Yang et~al.(2024)Yang, Teng, Zheng, Ding, Huang, Xu, Yang, Hong, Zhang, Feng, et~al.]{yang2024cogvideox}
Zhuoyi Yang, Jiayan Teng, Wendi Zheng, Ming Ding, Shiyu Huang, Jiazheng Xu, Yuanming Yang, Wenyi Hong, Xiaohan Zhang, Guanyu Feng, et~al.
\newblock Cogvideox: Text-to-video diffusion models with an expert transformer.
\newblock \emph{arXiv preprint arXiv:2408.06072}, 2024.

\bibitem[Yin et~al.(2024{\natexlab{a}})Yin, Gharbi, Park, Zhang, Shechtman, Durand, and Freeman]{yin2024improved}
Tianwei Yin, Micha{\"e}l Gharbi, Taesung Park, Richard Zhang, Eli Shechtman, Fredo Durand, and William~T Freeman.
\newblock Improved distribution matching distillation for fast image synthesis.
\newblock \emph{arXiv preprint arXiv:2405.14867}, 2024{\natexlab{a}}.

\bibitem[Yin et~al.(2024{\natexlab{b}})Yin, Gharbi, Zhang, Shechtman, Durand, Freeman, and Park]{yin2024one}
Tianwei Yin, Micha{\"e}l Gharbi, Richard Zhang, Eli Shechtman, Fredo Durand, William~T Freeman, and Taesung Park.
\newblock One-step diffusion with distribution matching distillation.
\newblock In \emph{Proceedings of the IEEE/CVF Conference on Computer Vision and Pattern Recognition}, pages 6613--6623, 2024{\natexlab{b}}.

\bibitem[Yin et~al.(2024{\natexlab{c}})Yin, Zhang, Zhang, Freeman, Durand, Shechtman, and Huang]{yin2024slow}
Tianwei Yin, Qiang Zhang, Richard Zhang, William~T Freeman, Fredo Durand, Eli Shechtman, and Xun Huang.
\newblock From slow bidirectional to fast causal video generators.
\newblock \emph{arXiv preprint arXiv:2412.07772}, 2024{\natexlab{c}}.

\bibitem[Zhai et~al.(2024)Zhai, Lin, Yang, Li, Wang, Lin, Doermann, Yuan, and Wang]{zhai2024motion}
Yuanhao Zhai, Kevin Lin, Zhengyuan Yang, Linjie Li, Jianfeng Wang, Chung-Ching Lin, David Doermann, Junsong Yuan, and Lijuan Wang.
\newblock Motion consistency model: Accelerating video diffusion with disentangled motion-appearance distillation.
\newblock \emph{arXiv preprint arXiv:2406.06890}, 2024.

\bibitem[Zhang et~al.(2025{\natexlab{a}})Zhang, Chen, Su, Ding, Stoica, Liu, and Zhang]{zhang2025fast}
Peiyuan Zhang, Yongqi Chen, Runlong Su, Hangliang Ding, Ion Stoica, Zhengzhong Liu, and Hao Zhang.
\newblock Fast video generation with sliding tile attention.
\newblock \emph{arXiv preprint arXiv:2502.04507}, 2025{\natexlab{a}}.

\bibitem[Zhang et~al.(2025{\natexlab{b}})Zhang, Huang, Chen, Lin, Liu, Stoica, Xing, and Zhang]{zhang2025faster}
Peiyuan Zhang, Haofeng Huang, Yongqi Chen, Will Lin, Zhengzhong Liu, Ion Stoica, Eric~P Xing, and Hao Zhang.
\newblock Faster video diffusion with trainable sparse attention.
\newblock \emph{arXiv e-prints}, pages arXiv--2505, 2025{\natexlab{b}}.

\bibitem[Zhang et~al.(2023)Zhang, Wei, Jiang, Zhang, Zuo, and Tian]{zhang2023controlvideo}
Yabo Zhang, Yuxiang Wei, Dongsheng Jiang, Xiaopeng Zhang, Wangmeng Zuo, and Qi Tian.
\newblock Controlvideo: Training-free controllable text-to-video generation.
\newblock \emph{arXiv preprint arXiv:2305.13077}, 2023.

\bibitem[Zhang et~al.(2025{\natexlab{c}})Zhang, Wei, Lin, Hui, Ren, Xie, and Zuo]{zhang2025videoelevator}
Yabo Zhang, Yuxiang Wei, Xianhui Lin, Zheng Hui, Peiran Ren, Xuansong Xie, and Wangmeng Zuo.
\newblock Videoelevator: Elevating video generation quality with versatile text-to-image diffusion models.
\newblock In \emph{Proceedings of the AAAI Conference on Artificial Intelligence}, pages 10266--10274, 2025{\natexlab{c}}.

\bibitem[Zhang et~al.(2025{\natexlab{d}})Zhang, Zhou, Zeng, Xu, Li, and Zuo]{zhang2025framepainter}
Yabo Zhang, Xinpeng Zhou, Yihan Zeng, Hang Xu, Hui Li, and Wangmeng Zuo.
\newblock Framepainter: Endowing interactive image editing with video diffusion priors.
\newblock \emph{arXiv preprint arXiv:2501.08225}, 2025{\natexlab{d}}.

\bibitem[Zhao et~al.(2024)Zhao, Jin, Wang, and You]{zhao2024real}
Xuanlei Zhao, Xiaolong Jin, Kai Wang, and Yang You.
\newblock Real-time video generation with pyramid attention broadcast.
\newblock \emph{arXiv preprint arXiv:2408.12588}, 2024.

\bibitem[Zheng et~al.(2023)Zheng, Nie, Vahdat, Azizzadenesheli, and Anandkumar]{zheng2023fast}
Hongkai Zheng, Weili Nie, Arash Vahdat, Kamyar Azizzadenesheli, and Anima Anandkumar.
\newblock Fast sampling of diffusion models via operator learning.
\newblock In \emph{International conference on machine learning}, pages 42390--42402. PMLR, 2023.

\bibitem[Zheng et~al.(2024)Zheng, Hu, Fan, Wang, Ding, Tao, and Cham]{zheng2024trajectory}
Jianbin Zheng, Minghui Hu, Zhongyi Fan, Chaoyue Wang, Changxing Ding, Dacheng Tao, and Tat-Jen Cham.
\newblock Trajectory consistency distillation.
\newblock \emph{arXiv preprint arXiv:2402.19159}, 2024.

\bibitem[Zhou et~al.(2024)Zhou, Zheng, Wang, Yin, and Huang]{zhou2024score}
Mingyuan Zhou, Huangjie Zheng, Zhendong Wang, Mingzhang Yin, and Hai Huang.
\newblock Score identity distillation: Exponentially fast distillation of pretrained diffusion models for one-step generation.
\newblock In \emph{Forty-first International Conference on Machine Learning}, 2024.

\end{thebibliography}
}

\clearpage
\setcounter{page}{1}
\maketitlesupplementary
\section{Further implementation details}
\noindent\textbf{Stage division and expert switching.}\quad During inference, we empirically observe that evenly dividing the total steps between the two experts produces favorable results. With 8 or 4 total steps, we assign 4 or 2 steps to each expert, respectively. These steps are uniformly sampled within each sub-trajectory.
\section{Additional Results}
\subsection{Compatibility with other acceleration techniques}
DCM accelerates generation via sampling step reduction and is compatible with other methods like low precision computation and sparse modeling. For example, integrating SVG~\cite{xi2025sparse} (which leverages the sparsity of 3D full attention), yields an additional 1.33× speedup on top of DCM-Hunyuan while maintaining high fidelity (VBench 83.79\%).

\subsection{Generality of DCM}
DCM addresses discrepancies in loss and gradient contributions across noise levels—a problem inherent to consistency distillation itself, not from any specific model architecture. Beyond HunyuanVideo~\cite{kong2024hunyuanvideo} and CogVideoX~\cite{yang2024cogvideox}, we further apply it to the recent WAN2.1-T2V~\cite{wan2025wan}. DCM significantly accelerates inference while preserving comparable visual quality, as evidenced by VBench scores (baseline: 83.2\%, DCM: 82.9\%).

\subsection{Visualization of the sampling process}
\begin{figure*}[t]
\centering
\includegraphics[width=.8\linewidth]{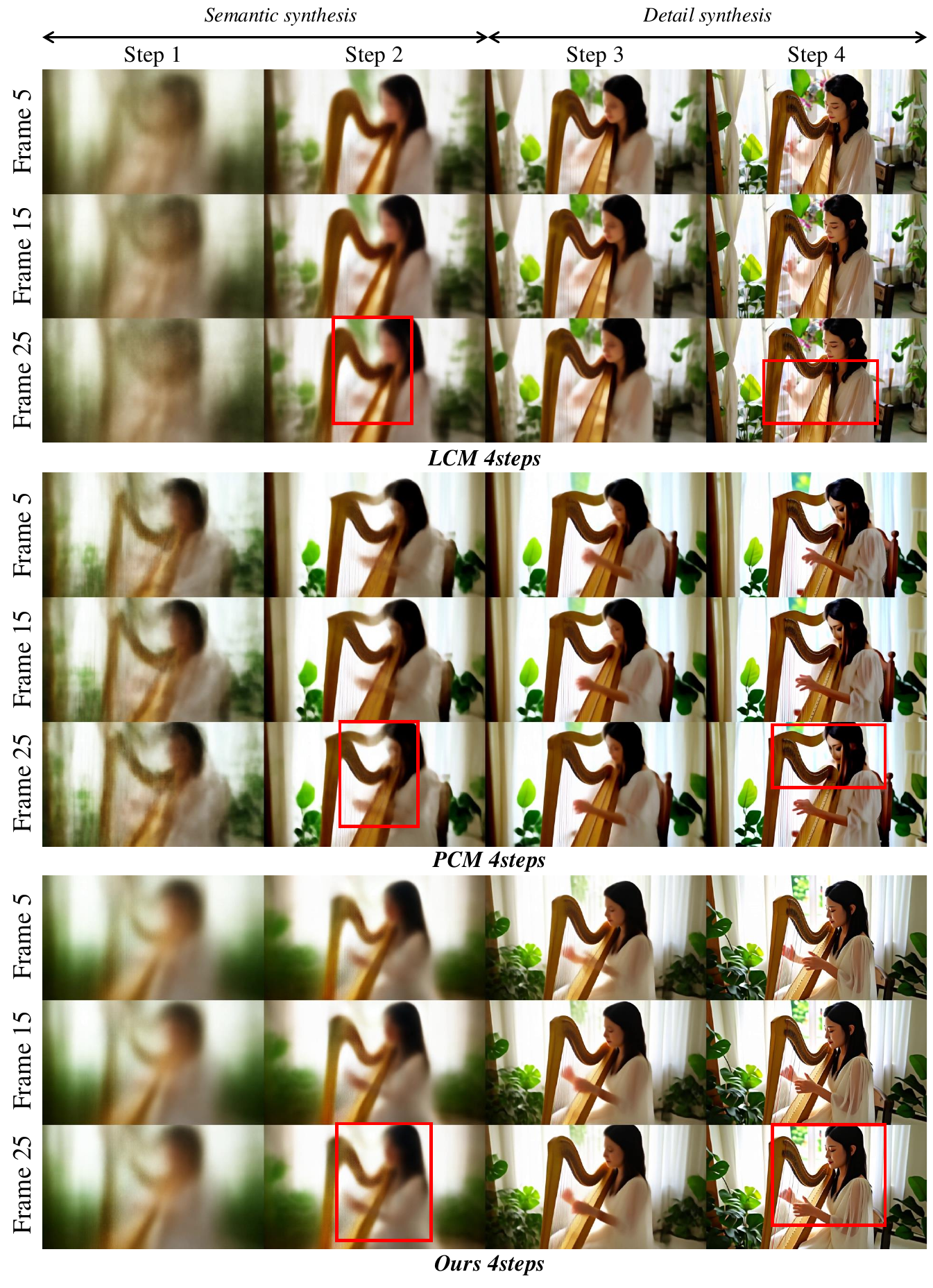}
\caption{Visualization of the sampling process of different methods.}
\label{fig:samproc}
\end{figure*}

\begin{figure*}[t]
\centering
\includegraphics[width=.8\linewidth]{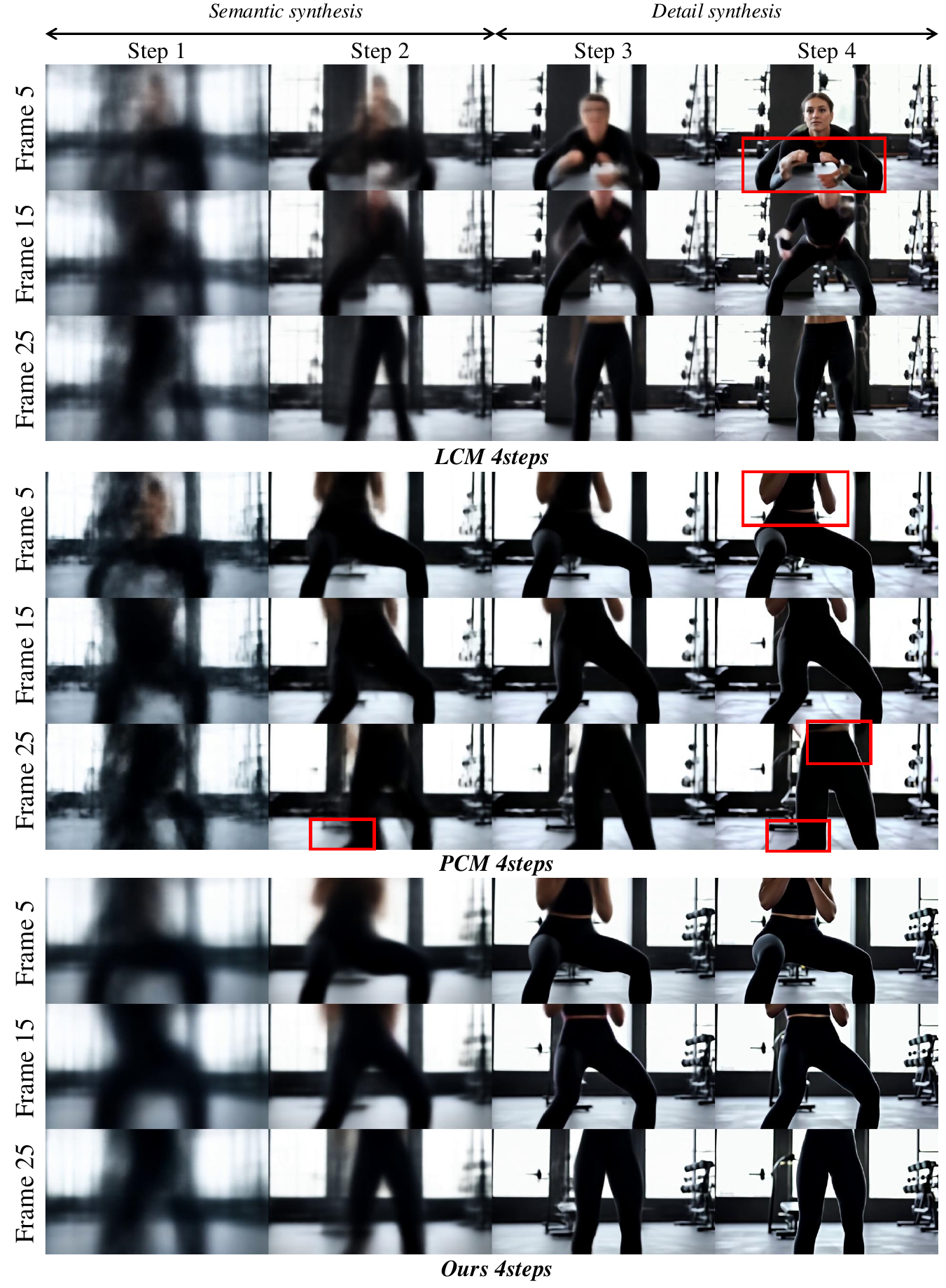}
\caption{Visualization of the sampling process of different methods.}
\label{fig:samprocB}
\end{figure*}

To further verify the effectiveness of our method in semantic and detail synthesis, we visualize the results of each sampling step in a 4-step sampling process on HunyuanVideo. As shown in Fig.~\ref{fig:samproc} and Fig.~\ref{fig:samprocB}, our method achieves better performance in both semantic layout and fine details compared to competing methods.

\subsection{More visual comparison results}
\begin{figure*}[t]
\centering
\includegraphics[width=.95\linewidth]{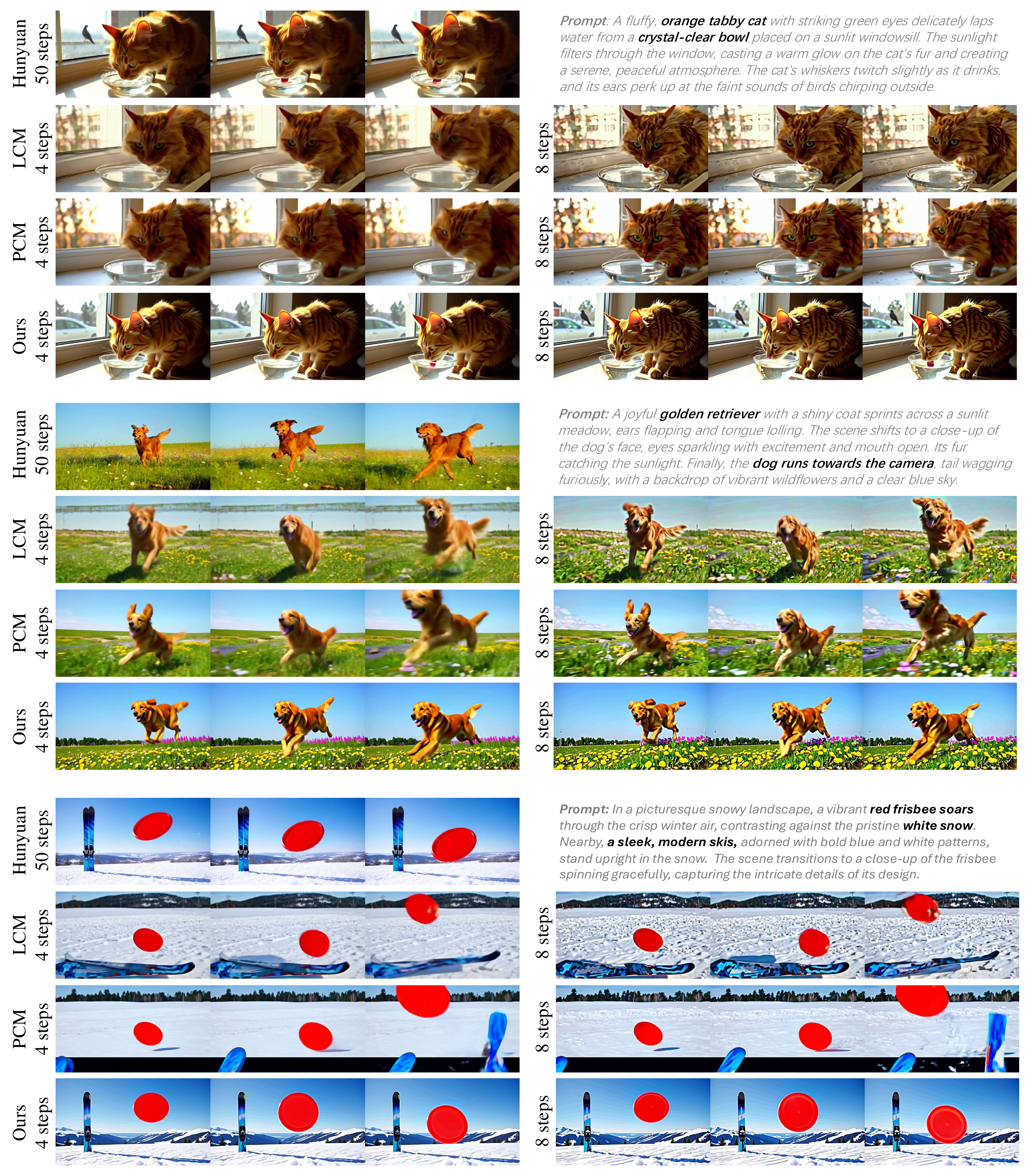}
\caption{Visual quality comparison of different methods.}
\label{fig:mhyA}
\end{figure*}

\begin{figure*}[t]
\centering
\includegraphics[width=.95\linewidth]{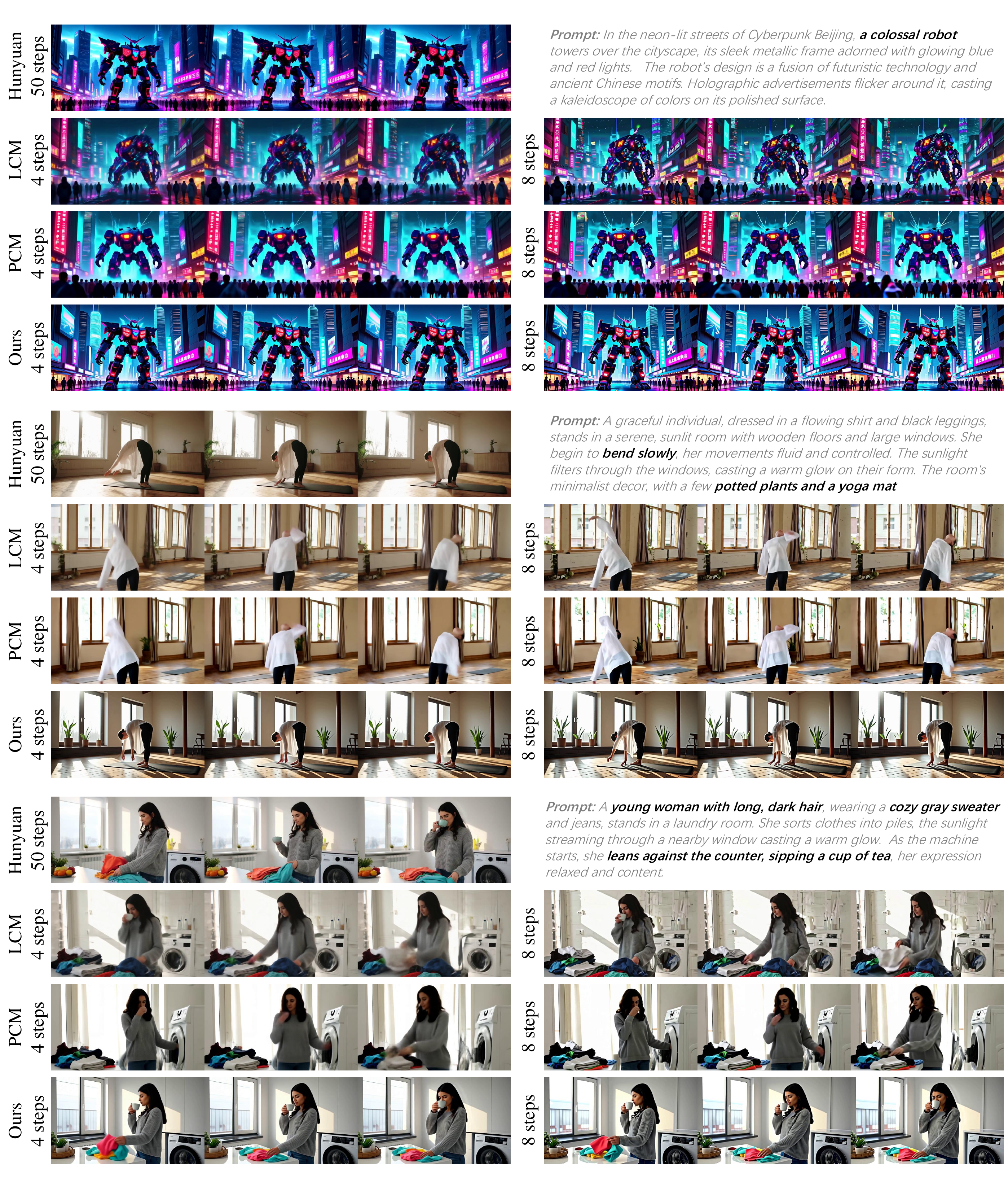}
\caption{Visual quality comparison of different methods.}
\label{fig:mhyB}
\end{figure*}

\begin{figure*}[t]
\centering
\includegraphics[width=.95\linewidth]{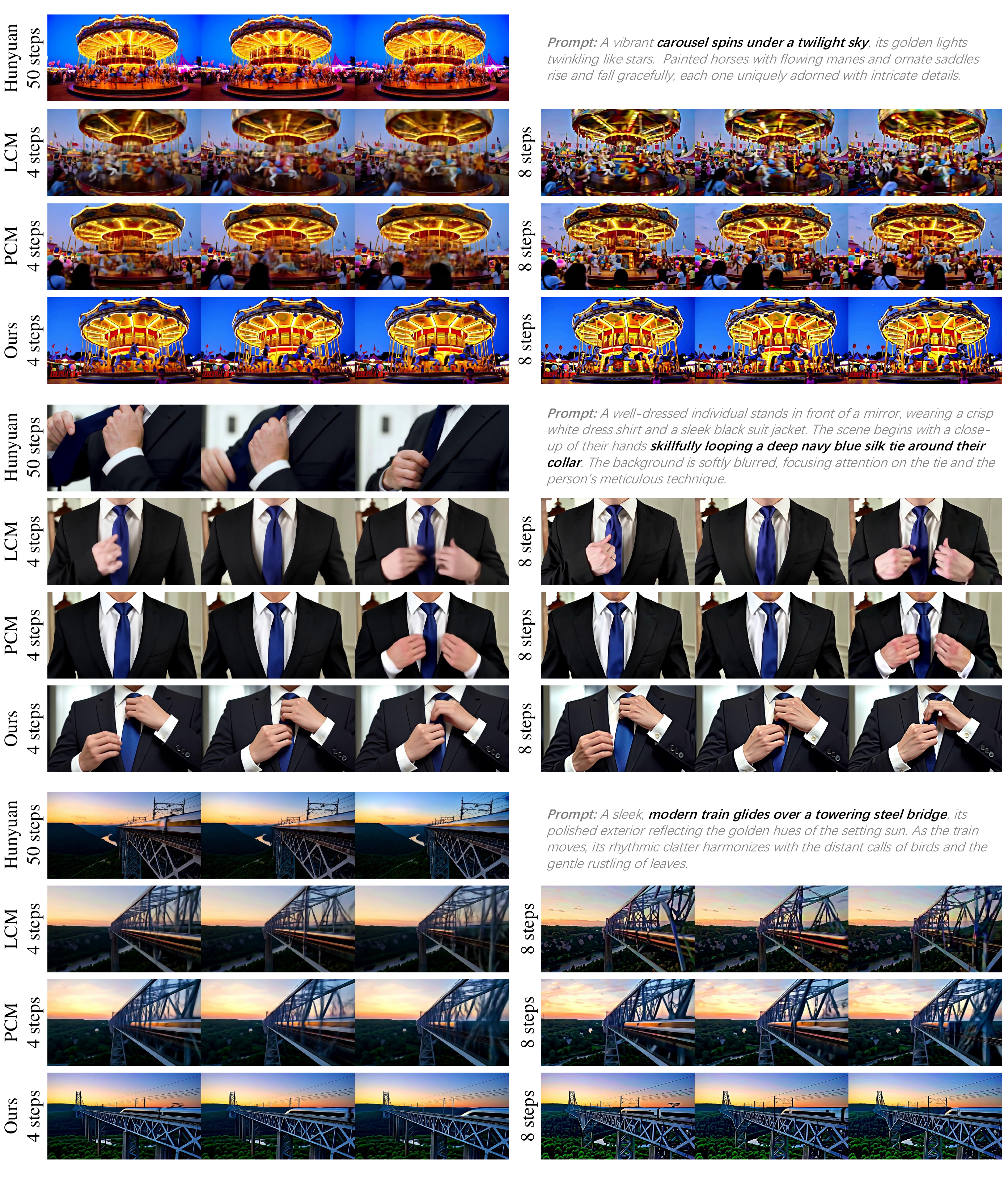}
\caption{Visual quality comparison of different methods.}
\label{fig:mhyC}
\end{figure*}

\begin{figure*}[t]
\centering
\includegraphics[width=.95\linewidth]{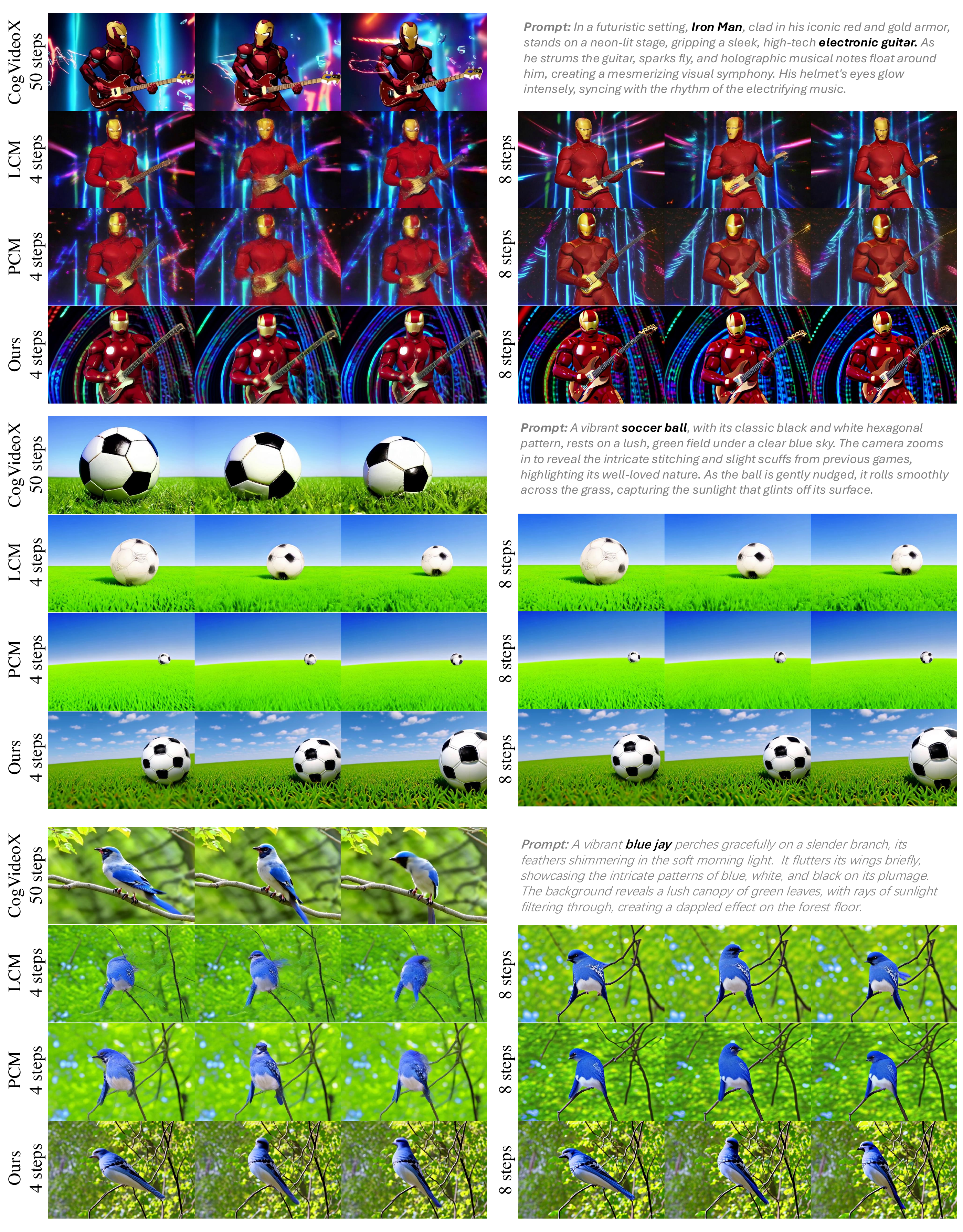}
\caption{Visual quality comparison of different methods.}
\label{fig:mcogA}
\end{figure*}

The additional visual comparison results for HunyuanVideo are presented in Fig.~\ref{fig:mhyA}, Fig.~\ref{fig:mhyB} and Fig.~\ref{fig:mhyC}. More visual results of CogVideoX are shown in Fig.~\ref{fig:mcogA}.
The results indicate that our method maintains reliable fidelity across diverse models, styles, and content in video synthesis while also achieving acceleration.

\end{document}